%% file: main.tex
\documentclass[11pt]{article}

\usepackage{acl}

\usepackage{times}
\usepackage{latexsym}

\usepackage[T1]{fontenc}

\usepackage[utf8]{inputenc}

\usepackage{microtype}

\usepackage{inconsolata}

\usepackage{graphicx}

\usepackage{hyperref}
\usepackage{url}
\usepackage{booktabs}       
\usepackage{amsfonts}       
\usepackage{nicefrac}       
\usepackage[table]{xcolor}         
\usepackage{amsmath}
\usepackage{amssymb}
\usepackage{mathtools}
\usepackage{amsthm}
\usepackage{bm}
\usepackage{booktabs}
\usepackage{multirow}
\usepackage[most]{tcolorbox}
\usepackage{enumitem}
\usepackage{subcaption}
\usepackage{rotating}
\usepackage[normalem]{ulem}
\usepackage{pifont}
\useunder{\uline}{\ul}{}
\usepackage{balance}

\title{Harnessing Consistency for Robust Test-Time LLM Ensemble}

\author{
 \textbf{Zhichen Zeng\textsuperscript{1}}\thanks{Equal contribution.},
 \textbf{Qi Yu\textsuperscript{1}}\footnotemark[1],
 \textbf{Xiao Lin\textsuperscript{1}},
 \textbf{Ruizhong Qiu\textsuperscript{1}},
 \textbf{Xuying Ning\textsuperscript{1}},\\
 \textbf{Tianxin Wei\textsuperscript{1}},
 \textbf{Yuchen Yan\textsuperscript{1}},
 \textbf{Jingrui He\textsuperscript{1}},
 \textbf{Hanghang Tong\textsuperscript{1}},
\\
\\
 \textsuperscript{1}University of Illinois Urbana-Champaign
\\
 \texttt{\{zhichenz,qiyu6,htong\}@illinois.edu}
}
\input{sections/hd}

\begin{document}
\maketitle

\input{sections/0-abs}
\input{sections/1-intro}
\input{sections/2-related}
\input{sections/3-method}
\input{sections/4-exp}
\input{sections/5-con}

\balance
\vspace{-10pt}
\bibliography{main}

\input{sections/app}

\end{document}

%% file: sections/hd.tex
\newcommand{\beqa}{\begin{eqnarray}}
\newcommand{\eeqa}{\end{eqnarray}}
\newcommand{\beq}{\begin{equation}}
\newcommand{\eeq}{\end{equation}}
\newcommand{\ben}{\begin{enumerate}}
\newcommand{\een}{\end{enumerate}}
\newcommand{\bit}{\begin{itemize}}
\newcommand{\eit}{\end{itemize}}
\newcommand{\bi}{\begin{itemize} \item}
\newcommand{\ei}{\end{itemize}}

\newtheorem{observation}{Observation}

\newcommand{\begindef}{\begin{Definition} \rm}
\newcommand{\beginexa}{\begin{Example} \rm}
\newcommand{\beginthe}{\begin{Theorem} \rm}
\newcommand{\beginpro}{\begin{Proposition} \rm}
\newcommand{\beginlem}{\begin{Lemma} \rm}
\newcommand{\begincon}{\begin{Conjecture} \rm}
\newcommand{\begincor}{\begin{Corollary} \rm}

\newcommand{\cmark}{\textcolor{teal}{\ding{51}}}
\newcommand{\xmark}{\textcolor{red}{\ding{55}}}
\newcommand{\codeinline}[1]{\colorbox{gray!15}{\texttt{#1}}}

\newcommand{\bluecell}[1]{\cellcolor[HTML]{CFE2F3}#1}
\newcommand{\yellowcell}[1]{\cellcolor[HTML]{FFF2CC}#1}
\newcommand{\redcell}[1]{\cellcolor[HTML]{FADBD8}#1}

\newcommand{\eat}[1]{}

\def\p{\bm{p}}
\def\tp{\tilde{\bm{p}}}
\def\A{\bm{A}}

\def\V{\mathcal{V}}

\def\tcs{\bm{s}^{\text{t}}}
\def\mcs{s^{\text{m}}}

\newtcolorbox{obsbox}[1]{
  colback=white!98!black,
  colframe=white!86!black,
  title={\textcolor{black}{#1}},
  boxrule=0.8pt,
  arc=2pt,
  left=6pt,
  right=6pt,
  top=0pt,
  bottom=0pt,
  before skip=5pt, 
}

\def\algname{\textsc{CoRE}}

%% file: sections/0-abs.tex
\begin{abstract}
    Different large language models (LLMs) exhibit diverse strengths and weaknesses, and LLM ensemble serves as a promising approach to integrate their complementary capabilities.
    Despite substantial progress in improving ensemble quality, limited attention has been paid to the robustness of ensembles against potential erroneous signals, which often arise from heterogeneous tokenization schemes and varying model expertise.
    Our analysis shows that ensemble failures typically arise from both the token level and the model level: the former reflects severe disagreement in token predictions, while the latter involves low confidence and pronounced disparities among models.
    In light of this, we propose \algname, a plug-and-play technique that harnesses model {\ul co}nsistency for {\ul r}obust LLM {\ul e}nsemble, which can be seamlessly integrated with diverse ensemble methods.
    \emph{Token-level consistency} captures fine-grained disagreements by applying a low-pass filter to downweight uncertain tokens with high inconsistency, often due to token misalignment, thereby improving robustness at a granular level.
    \emph{Model-level consistency} models global agreement by promoting model outputs with high self-confidence and minimal divergence from others, enhancing robustness at a coarser level.
    Extensive experiments across diverse benchmarks, model combinations, and ensemble strategies demonstrate that \algname\ consistently improves ensemble performance and robustness.
    Our code is available at \url{https://github.com/zhichenz98/CoRE-EACL26}.
\end{abstract}

%% file: sections/1-intro.tex
\section{Introduction}
Large language models (LLMs)~\cite{brown2020language,team2023gemini,touvron2023llama,achiam2023gpt,guo2025deepseek} have demonstrated remarkable performance in natural language processing tasks.
Due to the difference in model architectures, training algorithms, and datasets, different LLMs expertize in different areas, and it is important to ensemble various LLMs to integrate their complementary knowledge~\cite{yao2024determine,abdulaalbalancing,huang2024ensemble}.

Extensive efforts on test-time LLM ensemble can be broadly categorized into two categories: token-level and response-level ensemble.
Token-level ensemble aligns and fuses the token probabilities of different LLMs at each decoding step~\cite{yu2024breaking,yao2024determine,huang2024ensemble,xu2024bridging}, enabling fine-grained real-time correction for each token generation.
Response-level ensemble offers a more coarse-level ensemble by selecting either a complete response~\cite{jiang2023llm,lv2024urg,tekin2024llm} or a span~\cite{xu2024hit,lv2024specfuse,liu2024cool} from candidate outputs.
Despite their success in improving ensemble quality, existing approaches largely overlook ensemble robustness against noisy or erroneous signals.
For instance, incorrect token alignments can result in faulty probability fusion, while errors in model predictions may further compromise the correctness of ensemble outputs.
Therefore, it is crucial for ensemble methods to detect and mitigate such potential errors during inference to ensure reliable performance.

\begin{figure}[t!]
    \centering
    \includegraphics[width=0.95\linewidth]{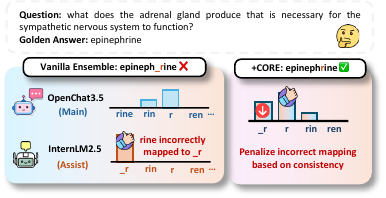}
    \caption{Motivation of \algname. \textbf{Left:} vanilla ensemble yields an incorrect prediction because the token \codeinline{\_r}, misaligned from the assistant token \codeinline{rine}, dominates the ensemble. \textbf{Right:} \algname\ penalizes inconsistent tokens, rendering correct prediction \codeinline{r}.}
    \vspace{-15pt}
    \label{fig:teaser}
\end{figure}

\begin{figure}[t]
    \centering
    \includegraphics[width=\linewidth]{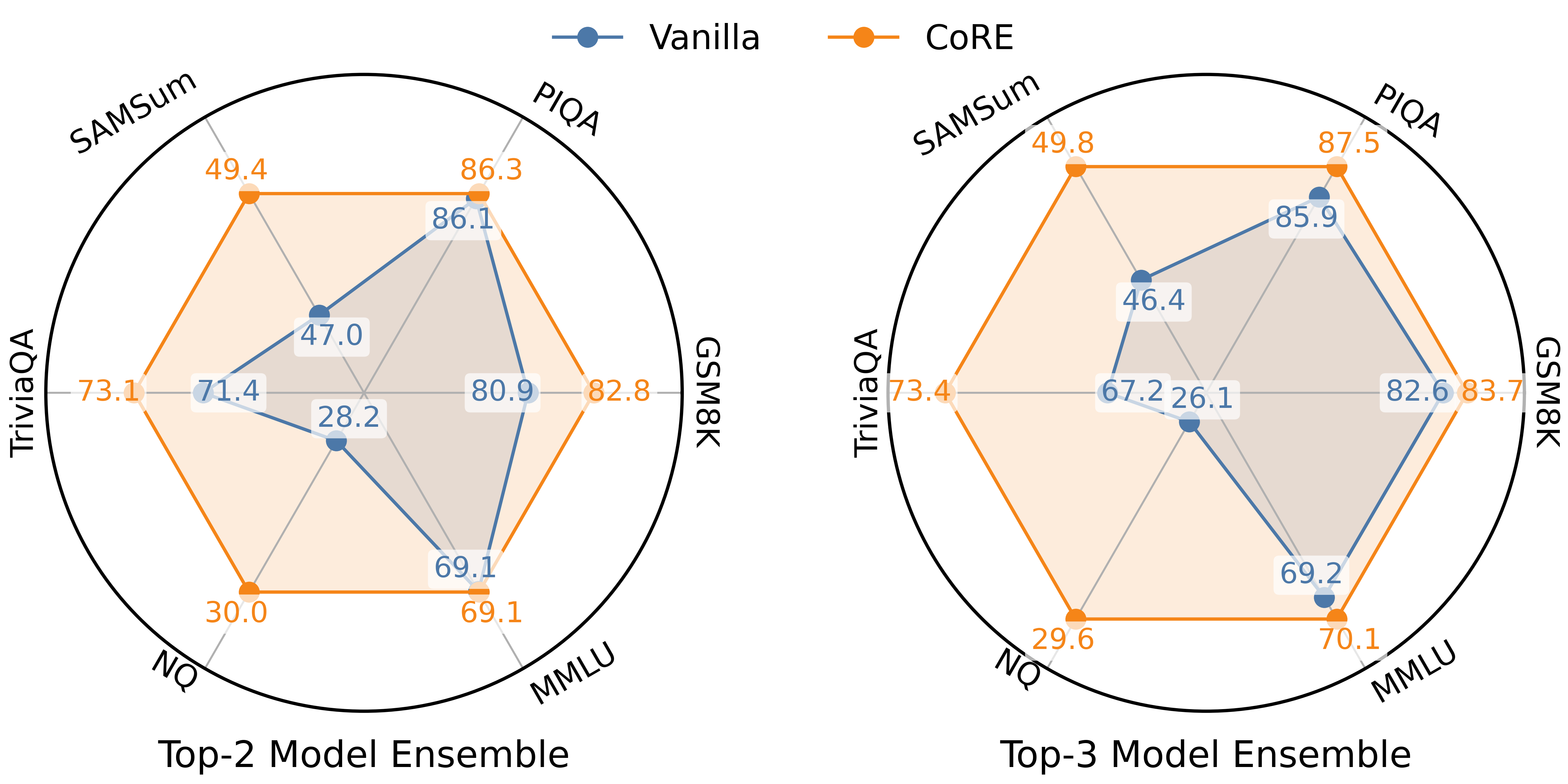}
    \caption{Ensemble performance across six benchmarks. We report the average performance of different ensemble methods with (\algname) and without \algname\  (Vanilla).}
    \vspace{-15pt}
    \label{fig:radar}
\end{figure}

To bridge this gap, we propose a plug-and-play technique named \algname\ to enhance both robustness and performance of LLM ensemble.
Our preliminary observations reveal that ensemble failures are closely tied to large discrepancies between the token distributions of different LLMs.
Motivated by this, we propose to harness both token and model consistency to achieve a robust ensemble.
At the token level, significant disparities in output probabilities for a specific token indicate fine-grained uncertainty, often stemming from token space misalignment. 
To address this, we introduce token consistency, which measures the disparity between each model's token probability and a reference probability, serving as a low-pass filter to amplify reliable tokens and suppress inconsistent ones.
At the model level, large divergence among full token distributions reflects model conflicts.
To capture this, we define model consistency that promotes model outputs with high self-confidence and low divergence from peers, thereby strengthening consistent models and downweighting unreliable ones.
A key advantage of the proposed \algname\ is that it is orthogonal to various token-level ensemble methods, and thus can be seamlessly integrated with \emph{no additional inference cost}.
We summarize the main contributions as follows.
\begin{itemize}[noitemsep, topsep=0pt]
    \item We systematically analyze and improve the robustness of LLM ensembles.
    \item We assess consistency at token and model levels to enhance ensemble performance and robustness. \algname\ can be seamlessly integrated with various ensemble methods to enable test-time correction with no additional cost.
    \item We conduct experiments across diverse benchmark tasks, model combinations, and ensemble methods. As shown in Figure~\ref{fig:radar}, \algname\ consistently improves baseline ensemble methods, achieving an average performance gain of 1.3\% and 2.8\% on Top-2 and Top-3 model ensemble, respectively.
\end{itemize}

%% file: sections/2-related.tex
\section{Related Works}

\subsection{Test-time LLM Ensemble}
Ensembling multiple large language models (LLMs) at test time offers a practical way to harness their diverse strengths and mitigate individual weaknesses. Existing methods can be broadly categorized into token and response level ensembles.

\textbf{Token‑level ensemble} \textit{fuses} next-token predictions across models at each decoding step, enabling fine-grained correction during generation.
Early works~\cite{fu2023specializing,wan2024knowledge,mavromatis2024pack} align token sequences via minimum edit distance, capturing structural difference but incurring high computational cost.
GAC~\cite{yu2024breaking} and UniTE~\cite{yao2024determine} bridge disparate vocabularies by aligning tokens through exact or prefix matches in text space.
Recent works, such as DeepEn~\cite{huang2024ensemble} and EVA~\cite{xu2024bridging}, learn projection functions that map heterogeneous token distributions into a shared representation space, enabling direct output fusion.

\textbf{Response-level Ensemble} \textit{selects} the most promising response from the multiple outputs to ensemble LLMs at a coarser-grained level.
One line of works select or synthesize a \emph{full response} among model outputs by either training-free approaches, such as perplexity scoring or majority voting~\cite{jiang2023llm,lv2024urg,tekin2024llm}, or training-based approaches that learn to rank responses~\cite{si2023getting}.
Another line of works~\cite{xu2024hit,lv2024specfuse,liu2024cool,cui2026adafuse} propose ensembling at the span level, aiming to strike a balance between fine-grained correction and context-aware decision-making.

\begin{figure*}[t!]
    \centering
    \includegraphics[width=\textwidth]{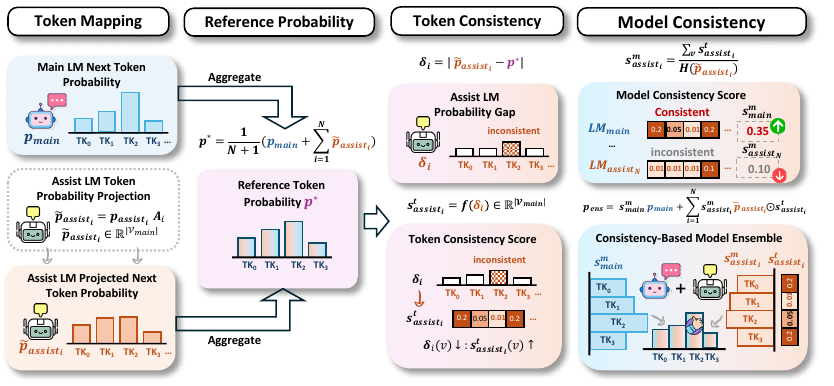}
    \caption{Overview of \algname. Token probabilities from different models are first aligned through token mapping, and a reference probability distribution is constructed for consistency evaluation. Token consistency mitigates the influence of inconsistent tokens, while model consistency highlights reliable and self-consistent models. Combining both yields a robust and improved ensemble prediction.}
    \vspace{-15pt}
    \label{fig:main}
\end{figure*}

\subsection{Model Consistency}
Recent research has explored consistency-based strategies to boost LLM performance by either  internal self-consistency or cross-model agreement.

\textbf{Self-Consistency} enhances answer reliability by aggregating diverse reasoning paths from a single model via frequency~\cite{wang2022self,li2024escape,aggarwal2023let}, entropy~\cite{kadavath2022language,lin2023generating,kang2025scalable}, or confidence signals~\cite{chen2023universal,taubenfeld2025confidence}, but it incurs notable computational overhead due to repeated sampling.

\textbf{Multi-model Consistency} combines outputs from different LLMs through majority voting~\cite{trad2024ensemble,niimi2025simple} or collaborative reasoning~\cite{wang2024mixture,liang2023encouraging,li2024improving}.
Despite the improved robustness, most focus on response-level agreement, overlooking fine-grained token-level consistency essential for detecting subtle errors in real time.

%% file: sections/3-method.tex
\section{Methodology}
In this section, we present our proposed \algname\ for a trustworthy LLM ensemble.
We begin by introducing the notation and formally defining the LLM ensemble problem in Section~\ref{sec:prob}. 
Then, in Section~\ref{sec:obs}, we reveal a strong correlation between ensemble performance and consistency scores at both the token and model levels. 
Motivated by these findings, we introduce our token-level consistency measure in Section~\ref{sec:token} and our distribution-level consistency measure in Section~\ref{sec:model}, both designed to improve ensemble performance and reliability.

\begin{figure*}[t]
    \centering
    \begin{subfigure}{.3\textwidth}
        \centering
        \includegraphics[width=\textwidth, trim=0 0 0 0, clip]{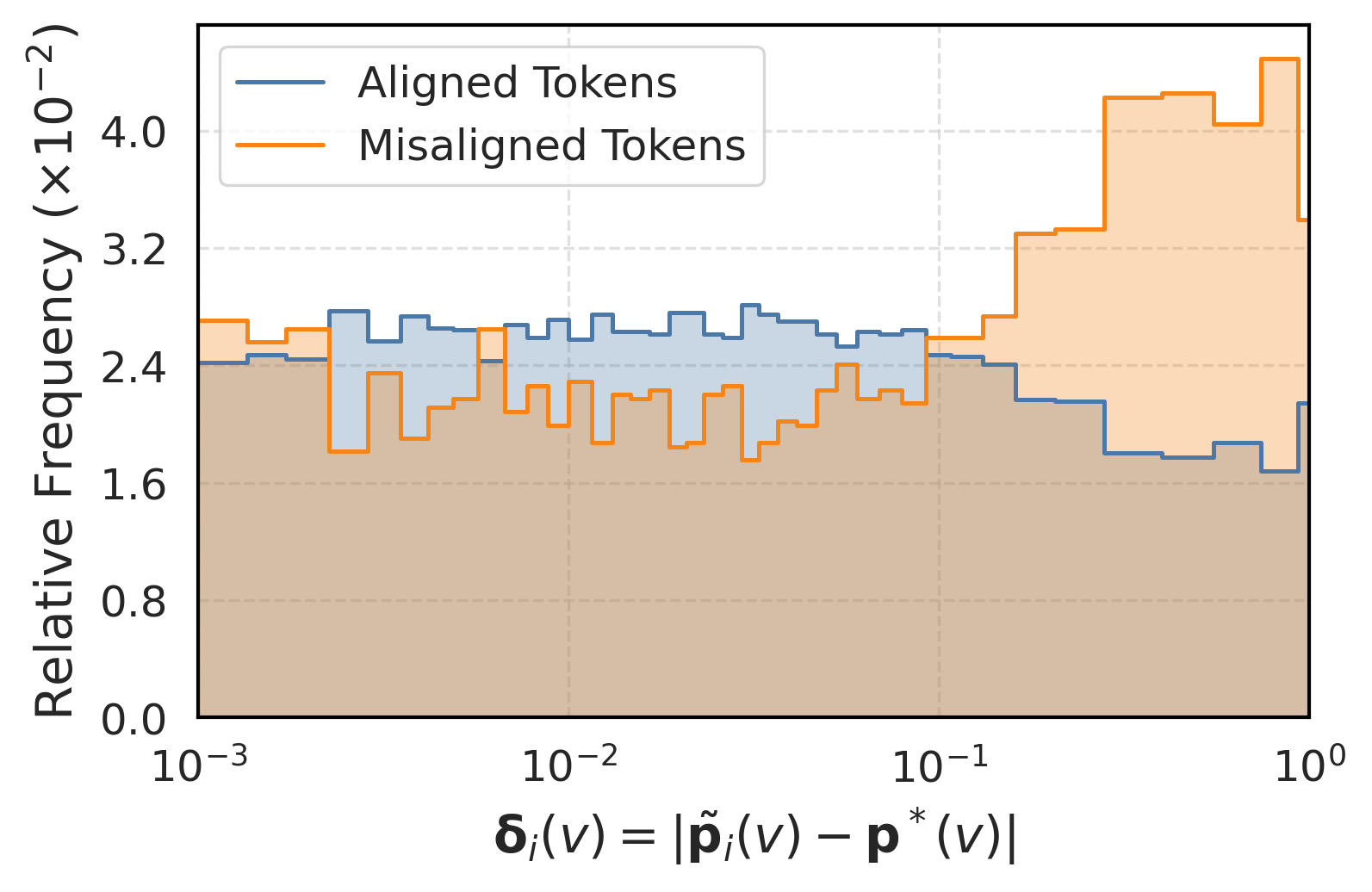}
        \caption{Token Probability Disparity.}
        \label{fig:obs-token}
    \end{subfigure}
    \hfill
    \begin{subfigure}{.3\textwidth}
        \centering
        \includegraphics[width=\textwidth, trim=0 0 0 0, clip]{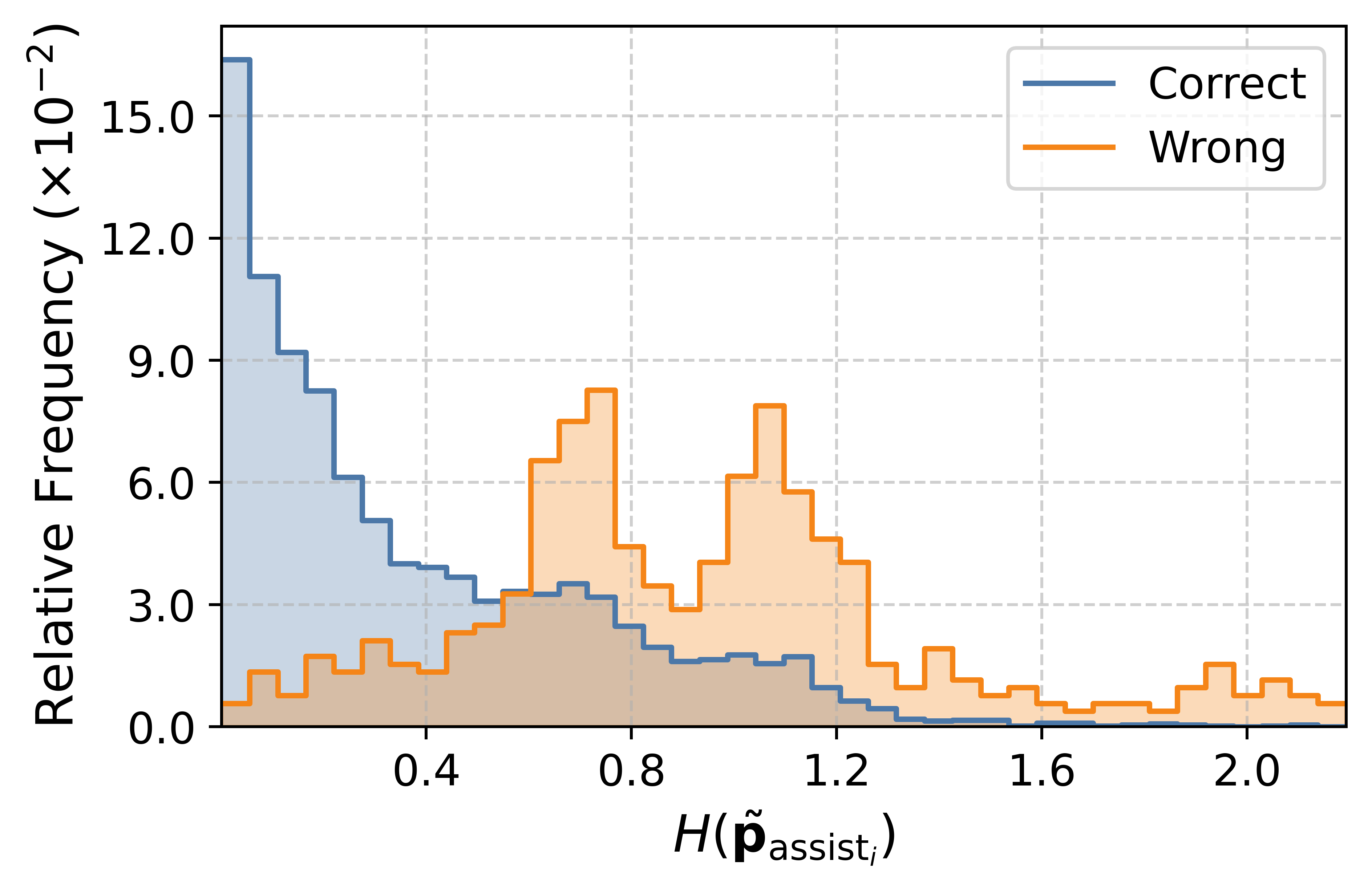}
        \caption{Entropy of Token Probability.}
        \label{fig:obs-model-conf}
    \end{subfigure}
    \hfill
    \begin{subfigure}{.31\textwidth}
        \centering
        \includegraphics[width=\textwidth, trim=0 8 0 0, clip]{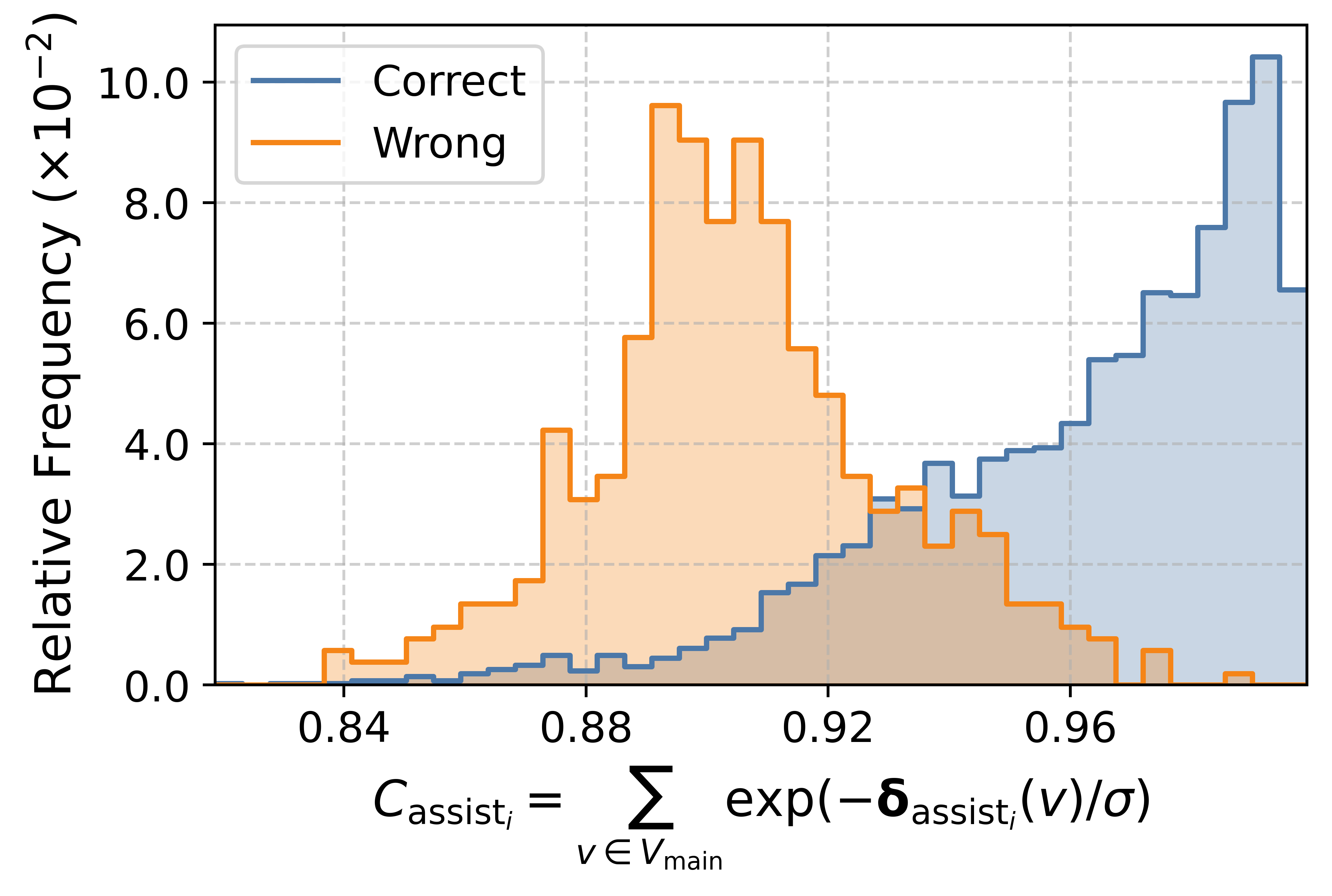}
        \caption{Sum of Token Consistency.}
        \label{fig:obs-model-cons}
    \end{subfigure}
    \vspace{-5pt}
    \caption{Observations. \textbf{(a) Token probability disparity:} Aligned tokens (blue) exhibit smaller disparities than misaligned ones (orange), indicating higher consistency. \textbf{(b) Entropy of token probability:} Correct answers (blue) exhibit lower entropy than wrong ones (orange), indicating higher confidence. \textbf{(c) Sum of token consistency:} Correct answers (blue) exhibit higher sum of token consistency score than wrong ones (orange).}
    \vspace{-10pt}
    \label{fig:obs}
\end{figure*}

\subsection{Problem Formulation}\label{sec:prob}
Formally, we are given a \emph{main model} with vocabulary set $\V_{\text{main}}$ and $N$ \emph{assistant models} with vocabulary set $\V_{\text{assist}_i}$ with $i\in\{1,2,...,N\}$.
We denote the predicted token distributions of main and assistant models as $\p_{\text{main}}$ and $\p_{\text{assist}_i}$, respectively.
LLM ensemble first learns a token alignment matrix $\A_{i}\in\mathbb{R}^{|\V_{\text{assist}_i}|\times |\V_\text{main}|}$ between the main model and the $i$-th assistant model, where $\A_i(v,u)=1$ indicates that token $v\in\V_{\text{assist}_i}$ is aligned with token $u\in\V_\text{main}$.
Token prediction probabilities can be further projected into the main model token space via $\p_{\text{assist}_i}\A_i\in\mathbb{R}^{|\V_{\text{main}}|}$.
For simplicity, we denote the aligned probability as $\tp_{\text{assist}_i}=\p_{\text{assist}_i}\A_i$.
The aligned probabilities can be further ensembled via
\begin{equation*}
    \p_{\text{ens}}=w_{\text{main}}\p_{\text{main}}+\sum_{i=1}^Nw_{\text{assist}_i}\tp_{\text{assist}_i}
\end{equation*}
where $\left[w_{\text{main}}, w_{\text{assist}_1}, ..., w_{\text{assist}_N}\right]\in\Delta^{N+1}$ are the normalized model weights.
For the special case where all models are assigned uniform weights, we denote the resulting ensembled probability distribution as $\p^*$, that is
\begin{equation}\label{eq:ref}
    \p^*=\frac{1}{N+1}\left(\p_{\text{main}}+\sum_{i=1}^N\tp_{\text{assist}_i}\right),
\end{equation}
In this paper, we use the above average probability as the reference $\p^*$ for consistency computation.

\subsection{Observations}\label{sec:obs}

We first study the distinct patterns in both token and model levels, which signal the inherent factors affecting the ensemble performance in Figure~\ref{fig:obs}.

First, we examine how token disparity, measure by the difference between the aligned probability and the reference probability on a specific token $v$, i.e., $\bm{\delta}_i(v) = |\tp_{\text{assist}_i}(v) - \p^*(v)|$, reflects the alignment correctness.
We experiment on 100 randomly-sampled questions from NQ dataset, and as shown in Figure~\ref{fig:obs-token}, aligned tokens concentrate at lower disparities, while misaligned tokens exhibit a heavier right tail.
A one-sided test of $H_0: \mathbb{E}_{v\in\text{misalign}}\bm{\delta}_i(v) \leq \mathbb{E}_{v\in\text{align}}\bm{\delta}_i(v)$ yields a $p$-value of $2.7\times 10^{-44}$, indicating that aligned tokens indeed have smaller disparity. We summarize this in Observation~\ref{obs:token}.
\begin{observation}[Token Consistency]\label{obs:token}
    Large token probability disparity signals token misalignment.
\end{observation}
For example, if token $\texttt{a}\in\V_{\text{assist}_i}$ is misaligned to token $\texttt{b}\in\V_{\text{main}}$, the difference in their probabilities $|\p_{\text{assist}_i}(\texttt{a})-\p^*(\texttt{b})|=|\tp_{\text{assist}_i}(\texttt{b})-\p^*(\texttt{b})|$ is expected to be large. 

Secondly, we evaluate how model confidence, measured by the entropy of the token probability $H(\tilde{\p}_{\text{assist}_i})$, relates to the answer correctness.
We experiment on PIQA dataset, and as shown in Figure~\ref{fig:obs-model-conf}, correct answers exhibit lower entropy than wrong ones.
A one-sided test of $H_0: \mathbb{E}[H(\tp)|\text{wrong}] \leq \mathbb{E}[H(\tp)|\text{correct}]$ yields a $p$-value of $7.7\times 10^{-108}$, confirming that wrong answers have larger entropy than correct ones.
We summarize this into Observation~\ref{obs:model-conf}.
\begin{observation}[Model Confidence]\label{obs:model-conf}
    Small model confidence measure by the entropy of token probability signals correct answer.
\end{observation}

Thirdly, we evaluate how model consistency, measured by the sum of RBF-transformed token disparities, i.e., $C_{\text{assist}_i}(v)=\allowbreak\sum_{v\in\V_{\text{main}}}\exp(-\bm{\delta}_{\text{assist}_i}(v)/\sigma)$, varies between correct and wrong answers.
We experiment on PIQA dataset, and as shown in Figure~\ref{fig:obs-model-cons}, correct answers exhibit higher model consistency than wrong ones.
A one-sided test of $H_0: \mathbb{E}[C_{\text{assist}_i} | \text{wrong}] \geq \mathbb{E}[C_{\text{assist}_i} | \text{correct}]$ yields a $p$-value of $5.0\times 10^{-222}$, indicating that correct answers have greater consistency.
We summarize this into Observation~\ref{obs:model-cons}.
\begin{observation}[Model Consistency]\label{obs:model-cons}
    Large model consistency measured by the sum of RBF-transformed token disparity signals correct answer.
\end{observation}

Motivated by these observations, we propose to harness both token and model consistency for robust LLM ensemble.

\subsection{Token Consistency}\label{sec:token}

We first introduce our proposed token consistency as a fine-grained measure.
A key challenge in LLM ensemble is the disparate token spaces due to different tokenization schemes adopted by different LLMs.
Despite extensive efforts in aligning token spaces, they inevitably encounter alignment errors that impair ensemble performance.

To address this limitation, it is essential to distinguish the misaligned tokens and rectify their contributions.
Motivated by Observation~\ref{obs:token}, where misaligned tokens exhibit large probability disparities, we propose token consistency as a low-pass filter to downweight their influence.
Adopting the average distribution in Eq.~\eqref{eq:ref} as the reference probability, we quantify the $i$-th assist model's token consistency $\tcs_{\text{assist}_i}$ as follows
\begin{equation}\label{eq:tcs}
    \begin{aligned}
        &\tcs_{\text{assist}_i}=f(\bm{\delta}_i)\in\mathbb{R}^{|\V_{\text{main}}|},\\
        &\text{where } \bm{\delta}_i=|\tp_{\text{assist}_i}-\p^*|\in\mathbb{R}^{|\V_{\text{main}}|},
    \end{aligned}
\end{equation}
where different functions, e.g., RBF kernel $f_{\text{rbf}}(\bm{\delta})=\exp(-\bm{\delta}/\sigma)$, power function $f_{\text{pow}}(\bm{\delta})=\alpha(1-\bm{\delta})^\beta$, and sigmoid function $f_{\text{sig}}(\bm{\delta})=1-\text{Sigmoid}(\gamma(\bm{\delta}_i-0.5))$, can be adopted.
Intuitively, large $\bm{\delta}$ induces small token consistency that signals large inconsistency with the reference probability.
By multiplying token consistency with the aligned distribution, i.e., $\tcs_{\text{assist}_i} \odot \tp_{\text{assist}_i}$, token consistency acts as a low-pass filter: penalizing inconsistent tokens with large disparities while promoting consistent tokens that are widely agreed upon.

\subsection{Model Consistency}\label{sec:model}
In addition to fine-grained token consistency, it is crucial to quantify the trustworthiness of the model.
Prior works typically assign model weights based on heuristics, such as uniform weighting~\cite{yao2024determine,yu2024breaking,xu2024bridging} or prior knowledge~\cite{huang2024ensemble}, or self-confidence metrics like perplexity~\cite{mavromatis2024pack,liu2024cool}.
However, the inter-model consistency remains largely underexplored.

Motivated by Observation~\ref{obs:model-conf} and \ref{obs:model-cons}, we argue that models exhibiting both high inter-model consistency and strong self-confidence should be prioritized.
To capture this, we define the model consistency by aggregating token consistency over the main token space, and regularizing it by the entropy of the output distribution serving as a proxy for confidence.
Formally, the model consistency of an assistant model is given by:
\begin{equation}\label{eq:mcs}
    \mcs_{\text{assist}_i}=\frac{\sum_{v\in\V_{\text{main}}}\tcs_{\text{assist}_i}(v)}{H(\tp_{\text{assist}_i})}\in\mathbb{R},
\end{equation}
where $H(\cdot)$ denotes the entropy of a distribution.
Here, the numerator rewards agreement with the reference model, while the denominator penalizes high uncertainty, thus favoring outputs that are both consistent and confident. 
A similar definition applies for the main model, denoted as $\mcs_{\text{main}}$.

By using token consistency as a token-wise filter and model consistency as model-level weights, the final ensembled distribution is computed as:
\begin{equation}
    \p_{\text{ens}} = \mcs_{\text{main}}\p_{\text{main}} + \sum_{i=1}^N \mcs_{\text{assist}_i}\tcs_{\text{assist}_i}\odot \tp_{\text{assist}_i},
\end{equation}
where $[\mcs_{\text{main}}, \mcs_{\text{assist}_1},\dots, \mcs_{\text{assist}_N}]\in\Delta^{N+1}$ are the normalized model consistency serving as the model weights.
Note that we apply token consistency only to assistant models to mitigate potential misalignment, as token misalignment arises solely when mapping tokens from assistant models to the 
main model's token space.

%% file: sections/4-exp.tex
\section{Experiments}\label{sec:exp}
We carry out extensive experiments to answer the following research questions:
\begin{itemize}[noitemsep, topsep=0pt]
    \item\textbf{RQ1:} How does \algname\ enhance ensemble performance? (Section~\ref{sec:exp-bench})
    \item\textbf{RQ2:} To what extent does \algname\ enhance ensemble robustness? (Section~\ref{sec:exp-robust})
    \item\textbf{RQ3:} What are the respective roles of token and model consistency, and how do their designs affect performance? (Section~\ref{sec:exp-study})
    \item\textbf{RQ4:} How does the performance scale with more models w/ and w/o \algname? (Section~\ref{sec:exp-study})
\end{itemize}
\subsection{Experiment Setup}\label{sec:exp-setup}
\paragraph{Base Models.}
We conduct our experiments on the following widely used models, including \texttt{Llama-3-8B-Instruct}~\cite{dubey2024llama}, \texttt{Mistral-7B-Instruct-v0.1}~\cite{jiang2023mistral7b}, \texttt{Qwen2.5-3b-Instruct}~\cite{team2024qwen2}, \texttt{InternLM2.5-7b-Chat}~\cite{team2023internlm} and \texttt{openchat-3.5-0106}~\cite{wang2023openchat}.

\paragraph{Ensemble Methods.}
We consider four baseline methods that align token spaces for LLM ensemble for benchmark evaluation, including:
\begin{itemize}[noitemsep, topsep=0pt]
    \item \textsc{MinED}~\cite{fu2023specializing,mavromatis2024pack} searches for textually closest tokens based on minimum edit distance.
    \item \textsc{GaC}~\cite{yu2024breaking} merges disparate token spaces into a union token space for ensemble;
    \item \textsc{UniTE}~\cite{yao2024determine} utilizes tokenizer to map tokens to their prefix counterparts;
    \item \textsc{EVA}~\cite{xu2024bridging} learns a mapping by aligning overlapping token embeddings.
\end{itemize}
Note that our proposed \algname\ mainly operates on the token space and is not specifically designed for methods like \textsc{DeePEn}~\cite{huang2024ensemble} that ensemble in the latent embedding space.
While being outside our main scope, we nonetheless conduct a separate analysis by slightly adapting \algname\ to integrate with \textsc{DeePEn}, in order to explore its potential benefits in this alternative setting.

\input{sections/tab_benchmark}

\input{sections/tab_base_model}

\paragraph{Datasets and Metrics.}
We evaluate six benchmarks covering four different categories, including: (1) \textbf{Reasoning:} \texttt{GSM8K}~\cite{cobbe2021training} (4-shot with CoT) covering grade school math problems, and \texttt{PIQA}~\cite{bisk2020piqa} (0-shot) with commonsense reasoning choice problems;
(2) \textbf{Summarization:} \texttt{SAMSum}~\cite{gliwa2019samsum} (0-shot) on dialogue summarization;
(3) \textbf{Knowledge:} \texttt{TriviaQA}~\cite{joshi2017triviaqa} (5-shot) and \texttt{NaturalQuestions (NQ)}~\cite{kwiatkowski2019natural} (5-shot);
(4) \textbf{Comprehensive Examination}: \texttt{MMLU}~\cite{hendrycks2009measuring} (5-shot) covering 57 subjects that humans typically learn.
We adopt Exact Match for \texttt{PIQA, TriviaQA, NQ} and \texttt{MMLU}, Accuracy for \texttt{GSM8K}, and Rouge-1 score for \texttt{SAMSum}.

\paragraph{Experiment Pipeline.}
We first evaluate the performance of base models on each dataset, and then select the Top-2 and Top-3 models for benchmark ensemble evaluation.
We adopt RBF function as the default consistency score for benchmark results, and set $\sigma=0.5$.
All experiments are conducted on NVIDIA A100 80GB GPUs.

\subsection{Benchmark Results}\label{sec:exp-bench}
We present the ensemble results in Table~\ref{tab:benchmark} and base models' performance in Table~\ref{tab:base}, from which we draw the following observations:

\noindent\textbf{(1) \algname\ achieves consistent improvements on different methods, datasets, and base model combinations.} 
Specifically, on reasoning datasets (\texttt{GSM8K}, \texttt{PIQA}), \algname\ enhances vanilla methods by 1.01 on Top-2 and 1.33 on Top-3 ensemble.
On the summarization dataset (\texttt{SAMSum}), \algname\ improves vanilla baselines by an average of 2.35 and 3.42 on Top-2 and Top-3 ensembles, respectively.
For knowledge-intensive datasets (\texttt{TriviaQA, NQ}), it yields average gains of 1.75 (Top-2) and 4.90 (Top-3).
On the comprehensive exam benchmark (\texttt{MMLU}), \algname\ provides smaller but consistent improvements of 0.03 (Top-2) and 0.94 (Top-3).

\noindent\textbf{(2) \algname\ achieves more stable ensemble.}
As more LLMs are included, baseline ensembles may suffer from negative ensemble, where performance degrades compared to the best single model.
In contrast, augmenting with \algname\ leads to robust and consistent improvements.
Notably, \algname\ successfully mitigates 17 negative ensemble cases encountered by the baseline ensemble methods.

Though not directly compatible with \textsc{DeePEn}, which performs ensemble in the latent embedding space rather than the token space, we adapt \algname\ to compute consistency over token embeddings and report the results in Table~\ref{tab:benchmark-deepen}.
Though less pronounced than token-space baselines, augmenting with \algname\ still yields consistent improvements.
We attribute this to two main factors: (1) \textsc{DeePEn} constructs its latent space using overlapping tokens that are identical, which inherently reduces token misalignment, and (2) the learned latent space already promotes cross-model agreement by design.
Interestingly, though some vanilla baselines, e.g., \textsc{MinED} and \textsc{UniTE}, may underperform \textsc{DeePEn}, they achieve better performance when augmented with \algname, validating the effectiveness of \algname.

\input{sections/tab_benchmark_deepen}

\begin{figure*}[t]
    \centering
    \begin{subfigure}{.48\linewidth}
        \centering
        \includegraphics[width=\linewidth, trim=0 0 0 8, clip]{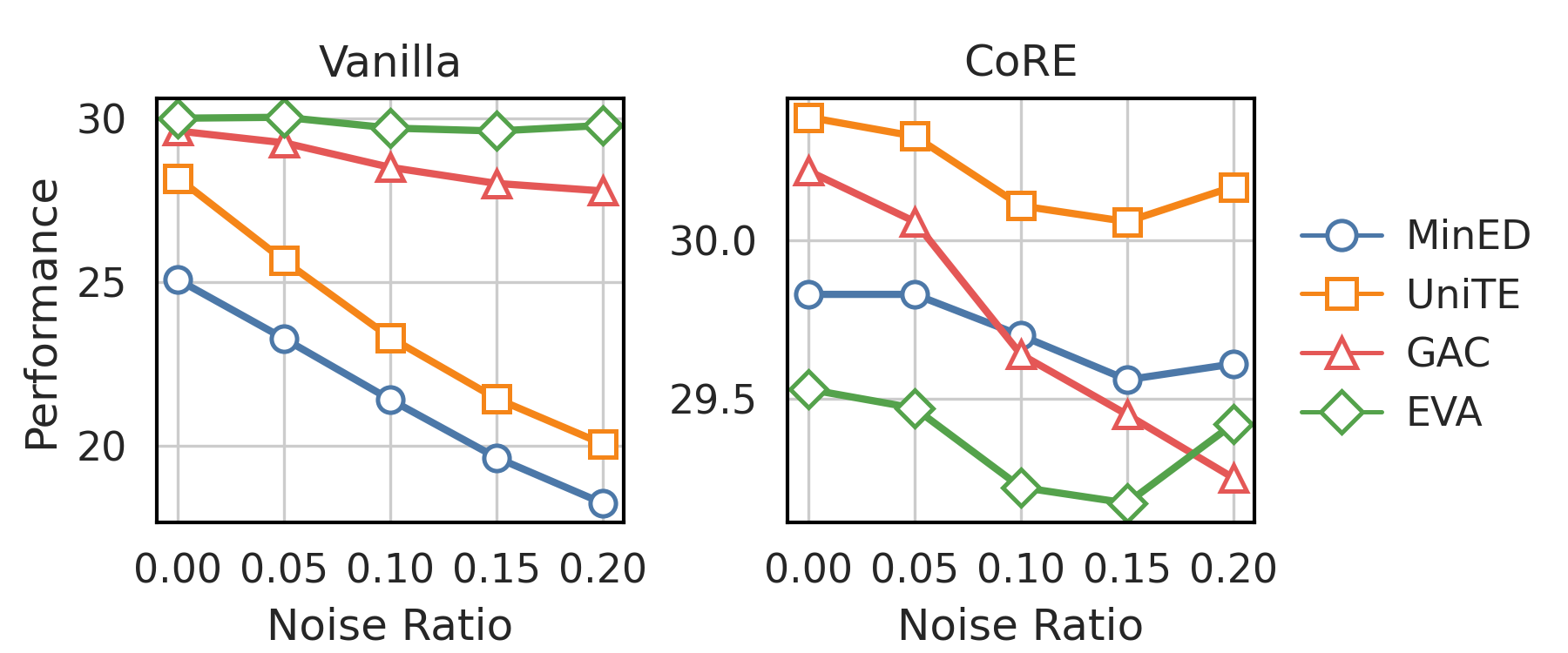}
        \vspace{-18pt}
        \caption{Token noises.}
        \label{fig:noise-token}
    \end{subfigure}
    \hfill
    \begin{subfigure}{.48\linewidth}
        \centering
        \includegraphics[width=\linewidth, trim=0 0 0 8, clip]{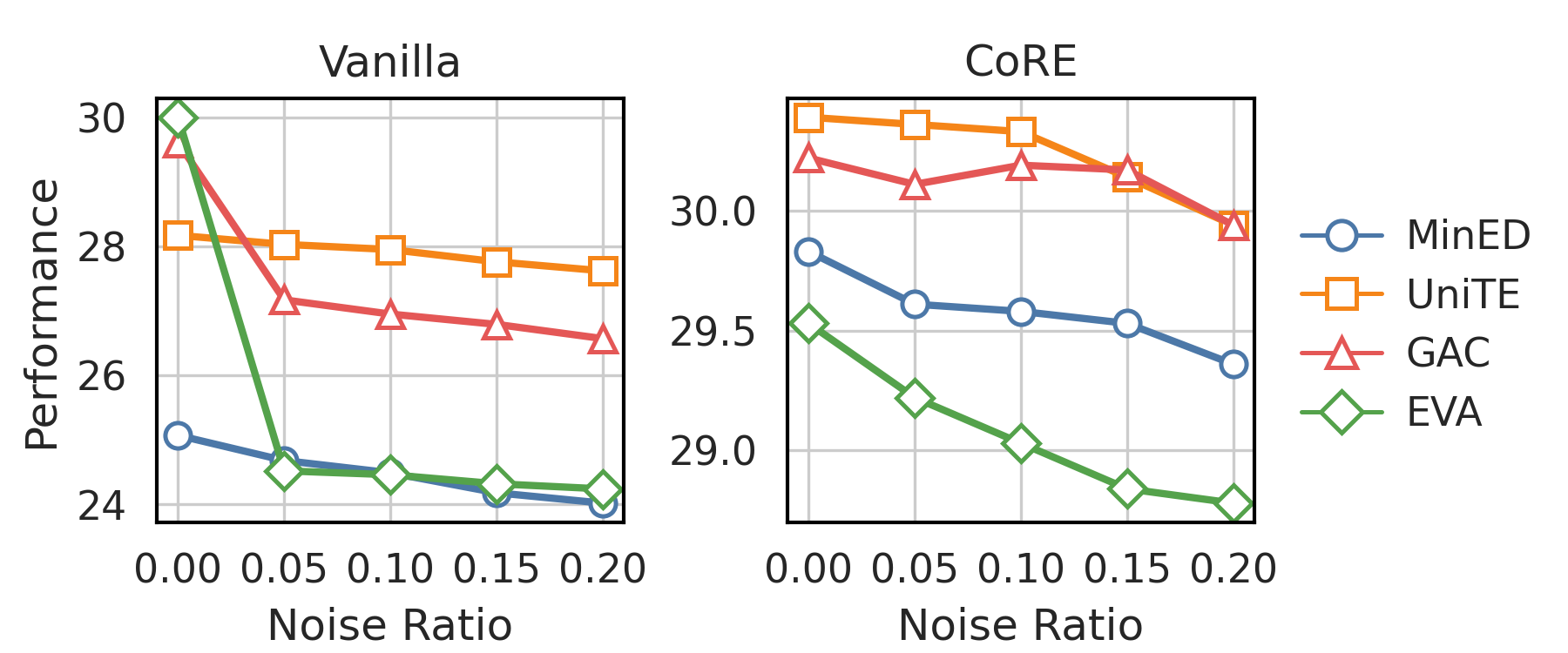}
        \vspace{-18pt}
        \caption{Probability noises.}
        \label{fig:noise-prob}
    \end{subfigure}
    \vspace{-10pt}
    \caption{Robustness against noises. \algname\ maintains stable performance with minimal degradation under noises.}
    \vspace{-10pt}
    \label{fig:noise}
\end{figure*}

\begin{figure}[t]
    \centering
    \includegraphics[width=\linewidth, trim=10 10 0 12, clip]{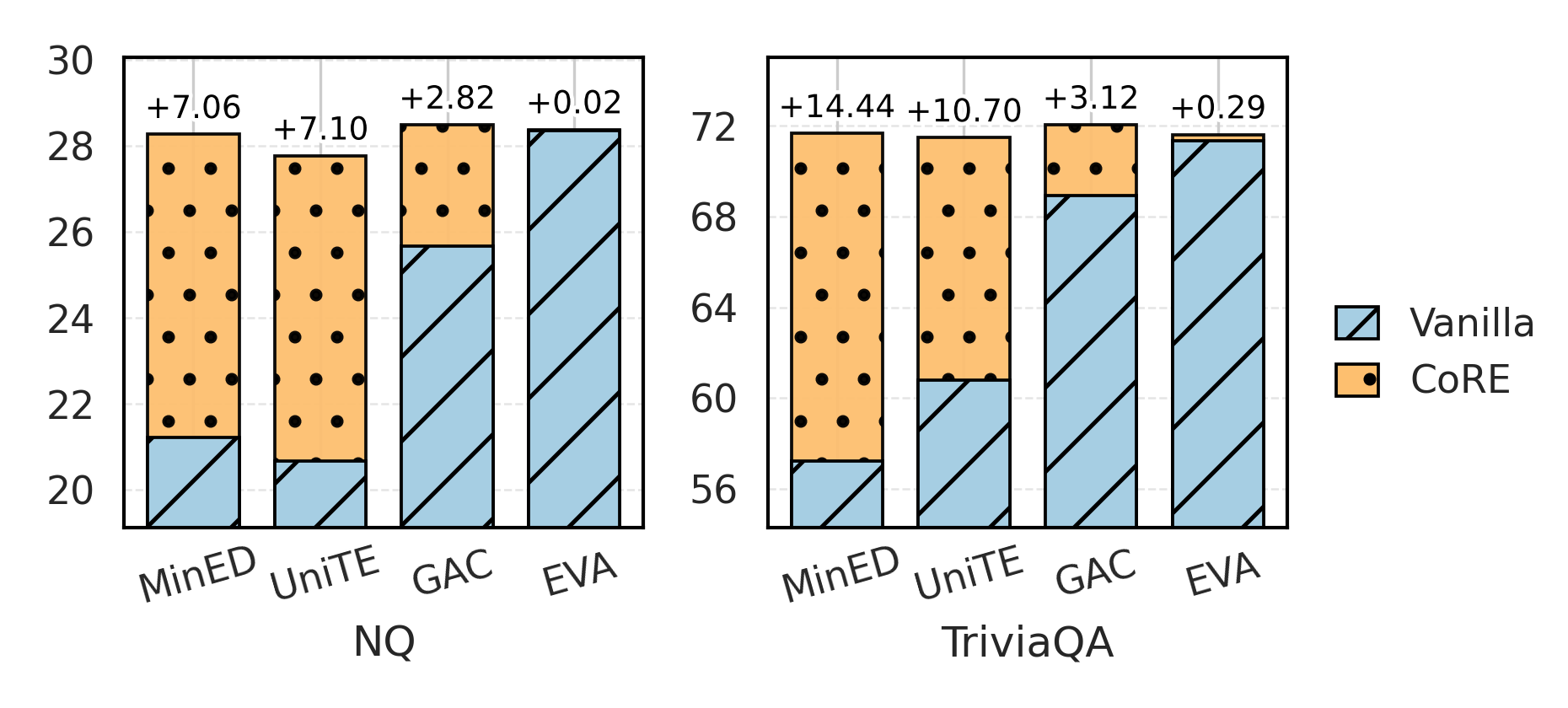}
    \vspace{-20pt}
    \caption{Robustness against large performance gap. We ensemble the best and worst performing models on \texttt{NQ} and \texttt{TriviaQA}. Numbers above the bars indicate the performance gains achieved by \algname.}
    \vspace{-15pt}
    \label{fig:exp-gap}
\end{figure}

\subsection{Robustness Results}\label{sec:exp-robust}

\subsubsection{Robustness against Noises}
We first evaluate ensemble performance against noises.
We consider two types of noises, including: (1) alignment noise, where 5\%, 10\%, 15\% and 20\% of the rows in the token mapping matrix are perturbed, and (2) probability noise, where Gaussian noise with standard deviations of 0.05, 0.10, 0.15, and 0.20 is added to the token probabilities.
The results are shown in Figures~\ref{fig:noise}.

Under both noise types, vanilla ensemble methods are highly sensitive, with average performance drops of 4.25 and 2.60 points as noise ratios increase from 0 to 0.2 for token and probability noise, respectively.
In contrast, \algname\ demonstrates strong robustness, exhibiting only marginal degradation of 0.38 and 0.49 under the same conditions.
Notably, \algname\ is particularly effective in mitigating the impact of token alignment noise.
This improvement can be attributed to the token consistency mechanism that identifies and rectifies misaligned tokens, thereby preserving ensemble accuracy even under severe perturbations.

\vspace{-3pt}
\subsubsection{Robustness against Performance Gap}
Previous methods~\cite{yao2024determine} may encounter negative ensemble, i.e., model performance drops after ensemble, when facing significant performance discrepancies among base models.
To evaluate \algname's robustness against performance gaps, we ensemble the best and worst performing models on \texttt{NQ} and \texttt{TriviaQA} with and without \algname, and the results are shown in Figure~\ref{fig:exp-gap}.

It is shown that \algname\ consistently improves all baseline ensemble methods on two datasets.
In particular, for vanilla methods that suffer severe degradation, e.g., \textsc{MinED}, \textsc{UniTE}, and \textsc{GaC}, augmenting with \algname\ yields substantial average gains of +5.66 on \texttt{NQ} and +9.42 on \texttt{TriviaQA}.
Moreover, while vanilla methods exhibit widely varying performance, their performance with \algname\ converges to comparable levels, highlighting \algname's ability to mitigate performance gaps and deliver stable improvements across diverse ensemble strategies.

\vspace{-5pt}
\subsection{Studies}\label{sec:exp-study}
\subsubsection{Scaling Results}
\begin{figure}[t]
    \centering
    \includegraphics[width=\linewidth, trim=0 0 0 8, clip]{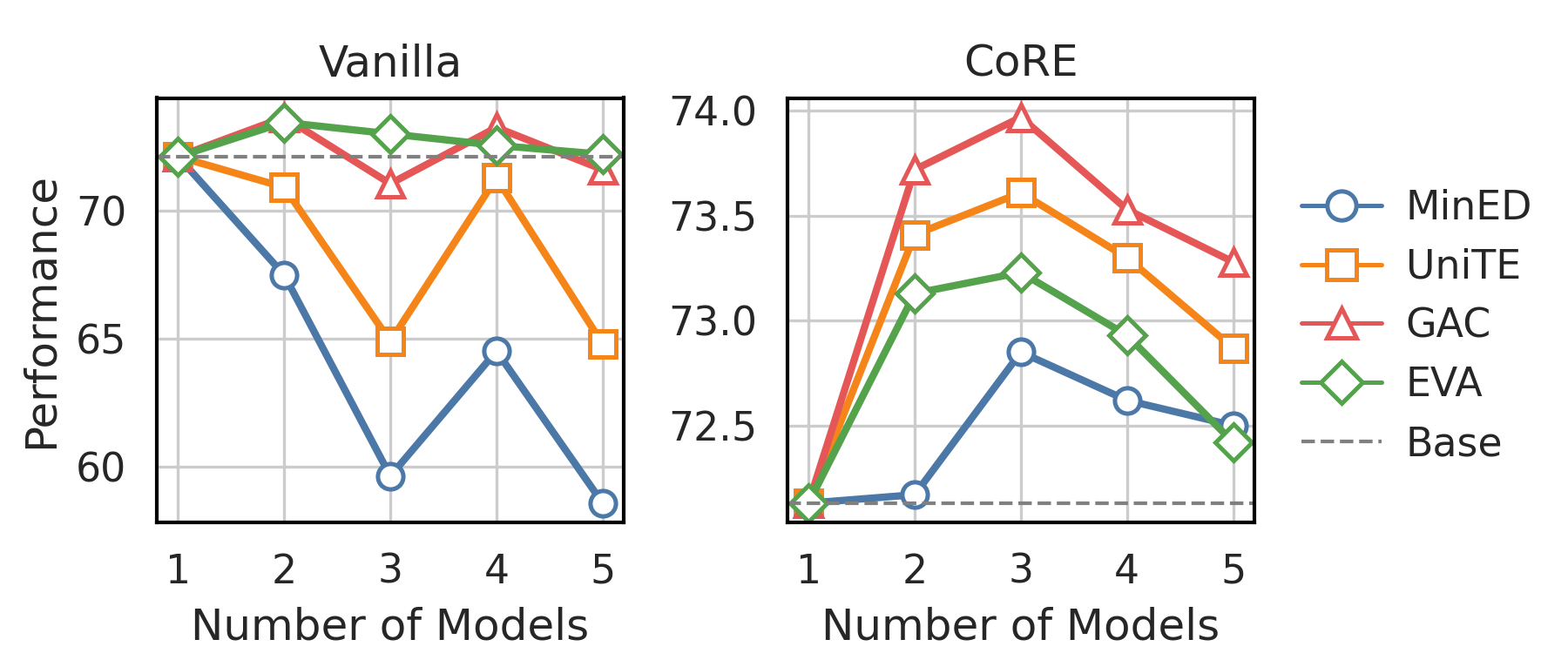}
    \vspace{-20pt}
    \caption{Scaling results on \texttt{TriviaQA}. \textbf{Left:} Vanilla ensemble methods suffer from negative ensembling, with performance degrading as more models are added. \textbf{Right:} \algname\ enables stable scaling, consistently outperforming the best single model across ensemble sizes.}
    \vspace{-15pt}
    \label{fig:exp-scale}
\end{figure}

We evaluate how \algname\ enhances ensemble scalability as more LLMs are incorporated.
As shown in Figure~\ref{fig:exp-scale}, vanilla ensembles suffer from negative ensembling, with performance often degrading as more models are added, whereas \algname\ enables stable scaling and consistently outperforms the best single model.
This performance degradation of the vanilla ensemble arises because increased model diversity introduces conflicting signals that, without consistency control, overwhelm the ensemble. 
Furthermore, token space alignment becomes more challenging when more LLMs are involved, and being unaware of the potential misalignment can lead to unstable and even counterproductive ensembling.
However, \algname\ enforces consistency and filters out unreliable outputs, transforming diversity into complementary information rather than noise, and thereby enabling robust and effective scaling.

\subsubsection{Ablation Study}
\begin{figure}[t]
    \centering
    \includegraphics[width=\linewidth, trim=10 10 20 12, clip]{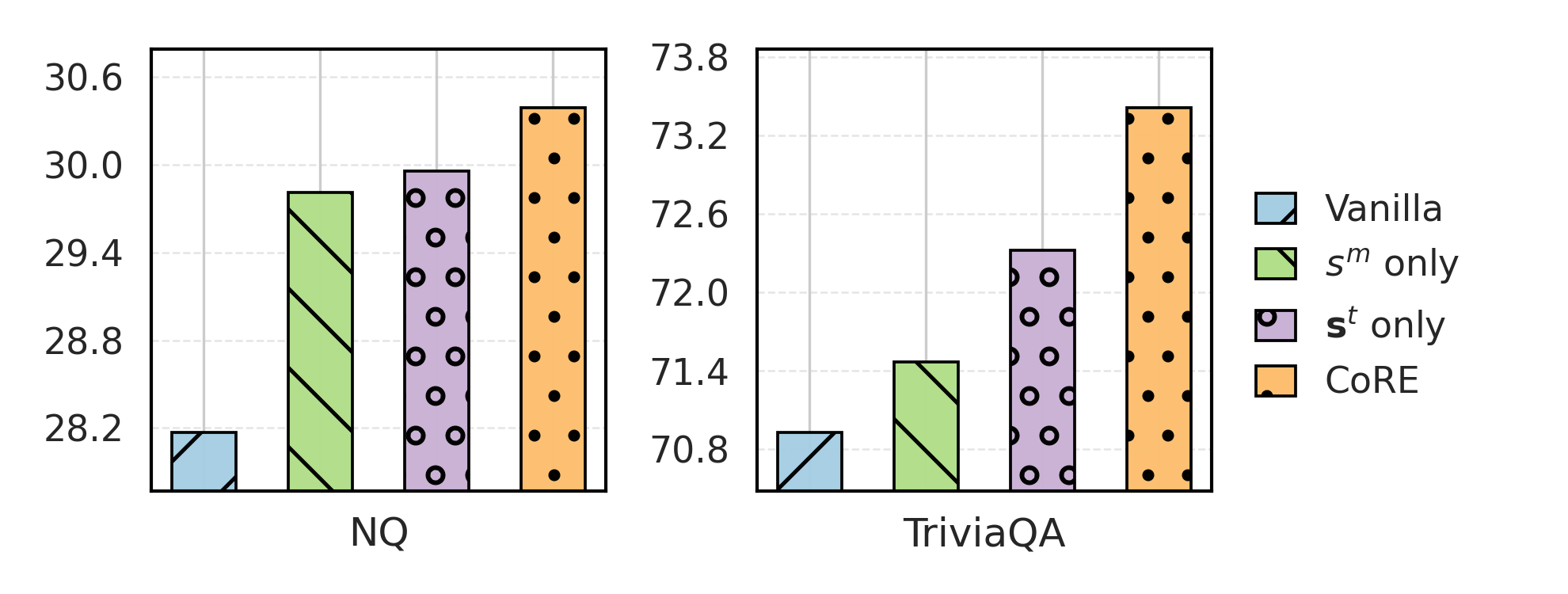}
    \vspace{-20pt}
    \caption{Ablation study. Both token and model consistency benefit the ensemble performance.} 
    \vspace{-15pt}
    \label{fig:ablate}
\end{figure}

\paragraph{On cosistency scores.}
We conduct ablation studies on token and model consistency on \texttt{NQ} and \texttt{TriviaQA}, and the results are shwon in Figure~\ref{fig:ablate}.
A clear trend emerges: \algname\ achieves the best performance, followed by applying only token consistency $\bm{s}^t$, then only model consistency $s^m$, while the vanilla ensemble performs the worst.
This indicates that both consistency components contribute positively to the performance: token consistency enhances fine-grained alignment across tokens, model consistency improves global agreement among models, and their joint integration yields the best overall gains.

\begin{figure}[h]
    \centering
    \includegraphics[width=\linewidth, trim=10 10 20 12, clip]{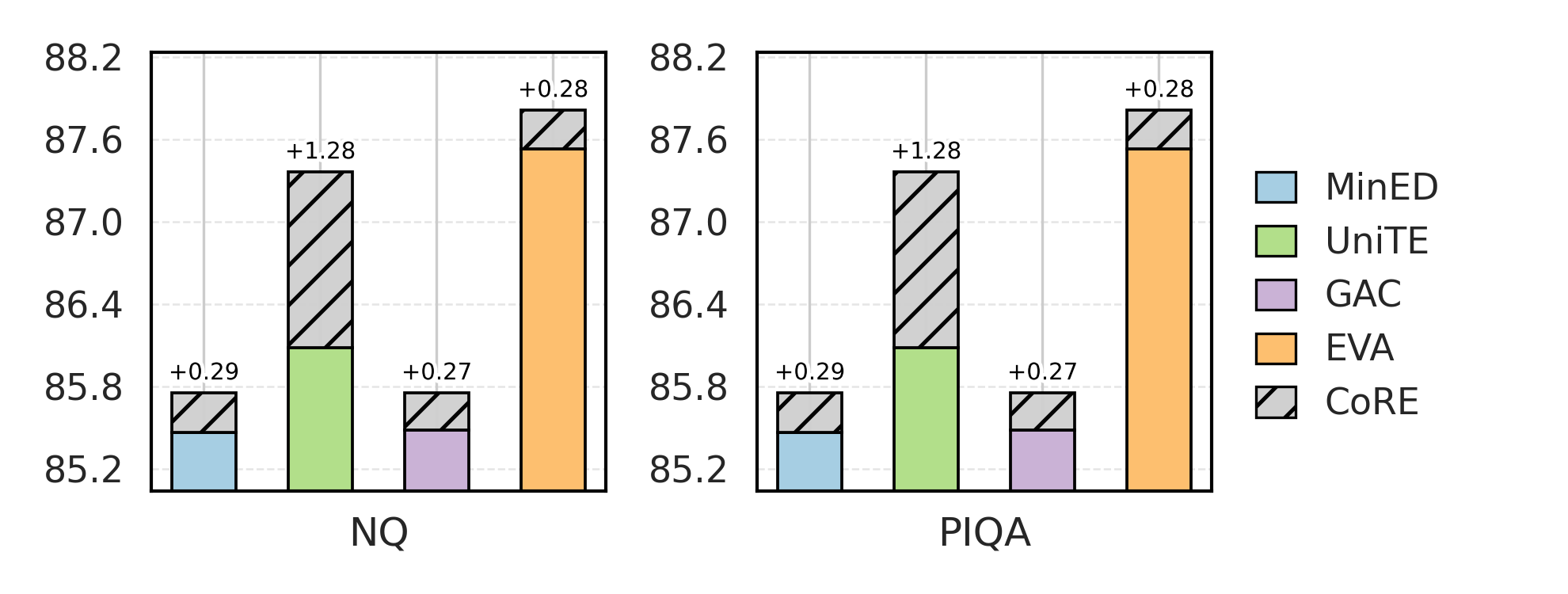}
    \caption{Ablation study of the entropy term. Colored bars indicate model consistency without the entropy term, while hatched bars indicate model consistency with the entropy term (\algname).}
    \label{fig:ablate_ent}
\end{figure}

\paragraph{On entropy weight.} 
We ablate the entropy term in the model consistency in Eq.~\eqref{eq:mcs}, and the results are presented in Figure~\ref{fig:ablate_ent}. 
From these results, we draw two key observations. 
First, ablating the entropy term consistently leads to performance degradation across all baselines. 
As shown by the hatched regions in the figure, reintroducing the entropy term yields performance gains ranging from +0.27 to +1.28, effectively validating its contribution to the model's overall accuracy. 
Notably, the UniTE model benefits most significantly, with a gain of +1.28. Second, even when the entropy term is ablated (colored bars), the performance remains robust compared to the vanilla ensemble, validating the effectiveness of the underlying consistency design independent of the entropy regularization.

\begin{figure}[h]
    \centering
    \includegraphics[width=\linewidth, trim=10 10 20 12, clip]{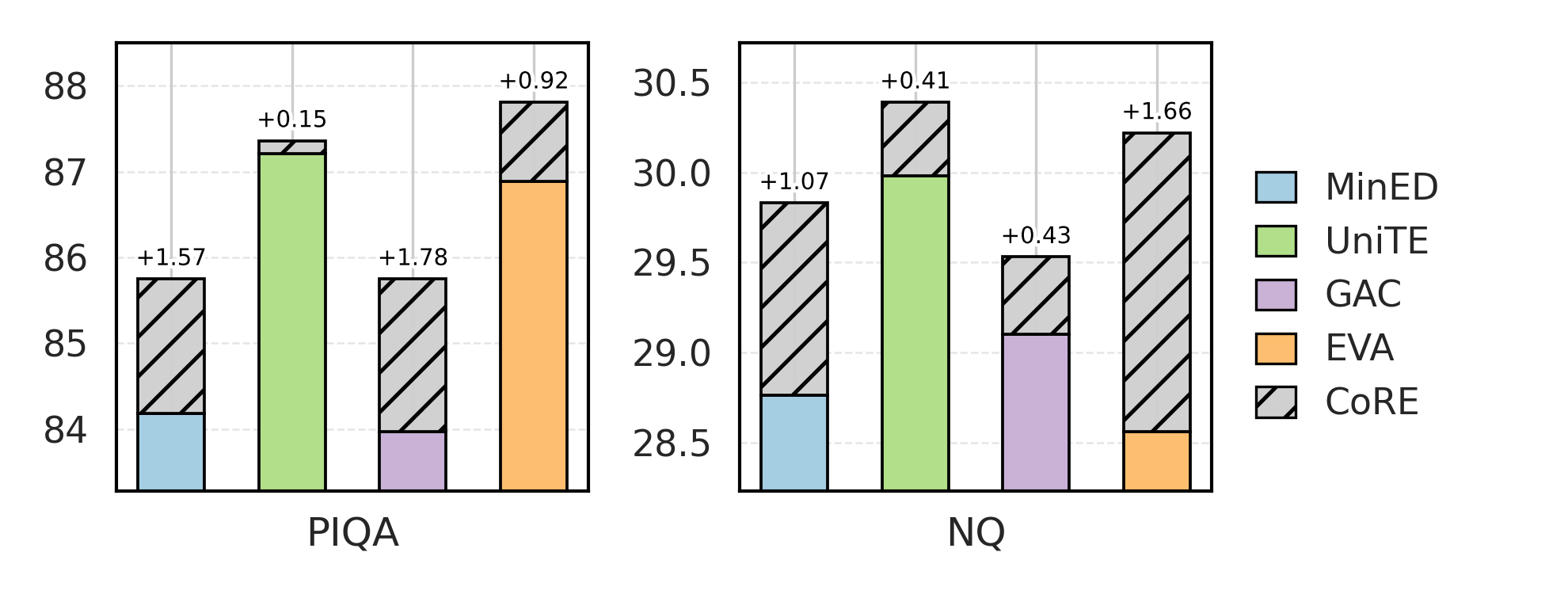}
    \caption{Ablation study of the token consistency. Colored bars indicate applying token consistency on both main and assist models, while hatched bars indicate only applying token consistency on assist models (\algname).}
    \vspace{-5pt}
    \label{fig:ablate_tc}
\end{figure}

\noindent\textbf{On token consistency design.}
The token consistency design in Eq.~\eqref{eq:tcs} is asymmetric, where only assist models are penalized while the main model remains original. 
To validate this design, we compare it against a symmetric variant where token consistency is applied to both main and assist models. 
The results are reported in Figure~\ref{fig:ablate_tc}.
We observe that the symmetric application (colored bars) consistently underperforms the proposed assistant-only design (hatched bars, \algname). 
As indicated by the annotated gains in the figure, removing the penalty from the main model leads to significant improvements, such as +1.78 on PIQA with GAC and +1.66 on NQ with EVA.
This confirms that the primary inconsistencies stem from assistant-to-main misalignment; therefore, penalizing the main model introduces unnecessary constraints that hurt the overall ensemble performance.

\subsection{Hyperparameter Study}
\begin{figure*}
    \centering
    \includegraphics[width=.8\linewidth]{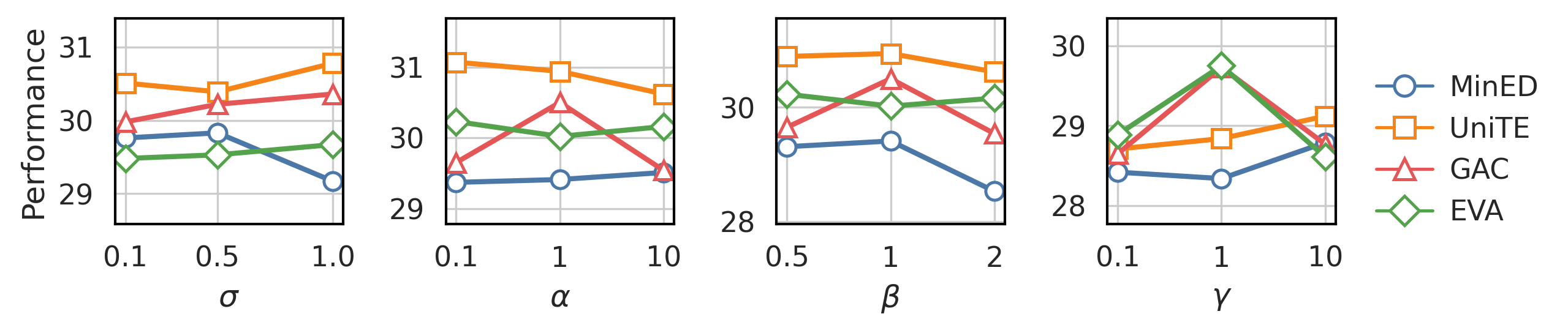}
    \vspace{-10pt}
    \caption{Hyperparameter study on $\sigma$ in RBF function, $\alpha$ and $\beta$ in power function, and $\gamma$ in sigmoid function.}
    \label{fig:hyper}
\end{figure*}
We conduct a sensitivity analysis on the hyperparameters governing the consistency functions: $\sigma$ in the RBF kernel, $\alpha$ and $\beta$ in the power function, and $\gamma$ in the sigmoid function.
As shown in Figure~\ref{fig:hyper}, across a broad range of values, \algname\ is fairly robust to these hyperparameters and does not require heavy fine-tuning.

Conceptually, these hyperparameters control the \textit{sharpness} of the consistency score, i.e., how aggressively inconsistent tokens and models are penalized.High values of $\alpha, \beta, \gamma$ (or small $\sigma$) yield "peaked" scores that strictly downweight inconsistencies, effectively filtering noise when assistant models are weak. Conversely, lower values (or large $\sigma$) produce smoother scores that tolerate more diversity, which is beneficial when the ensemble consists of strong, reliable models.

\subsubsection{Consistency Score Functions}
\begin{figure}[t]
    \centering
    \includegraphics[width=\linewidth, trim=10 15 0 10, clip]{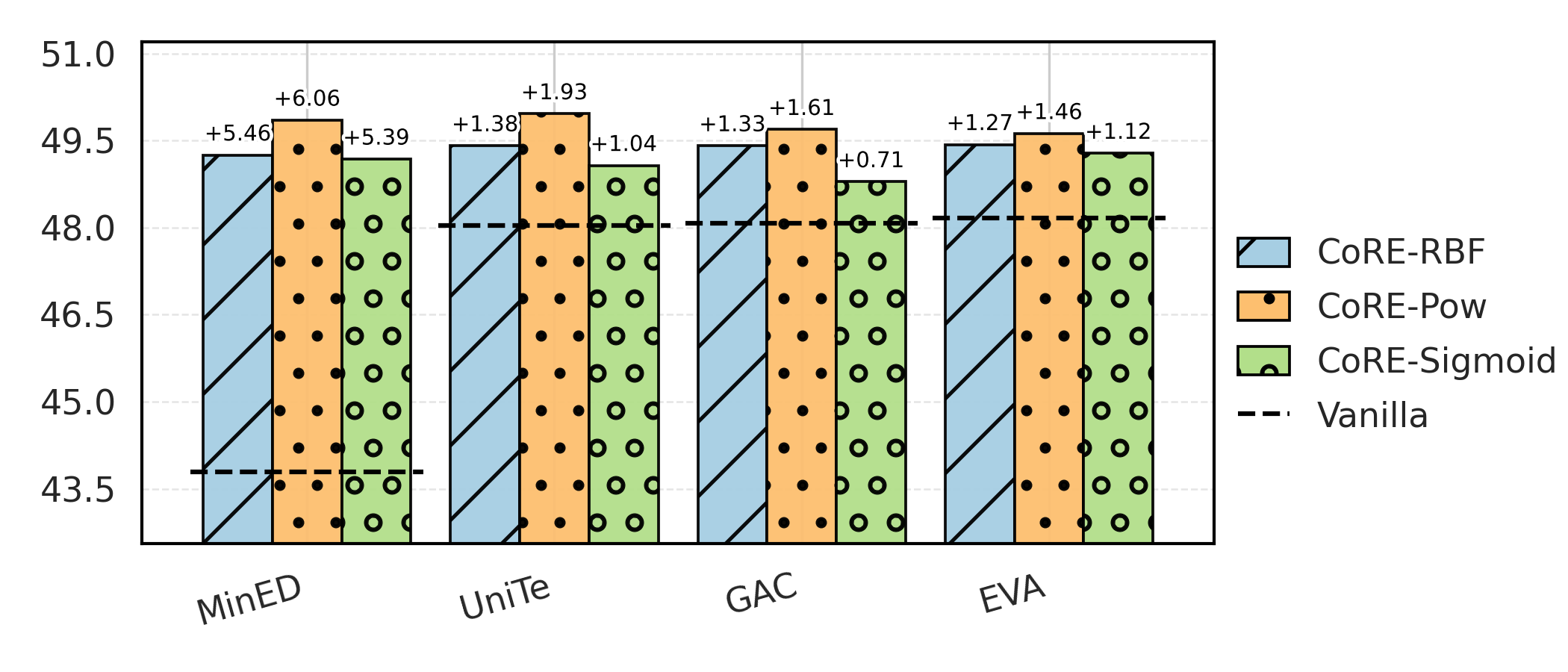}
    \vspace{-20pt}
    \caption{Ensemble performance with different consistency score functions on \texttt{SAMSum}. Numbers above the bars indicate the performance gains achieved.}
    \vspace{-10pt}
    \label{fig:exp-func}
\end{figure}

We explore alternative designs for consistency score to evaluate the generalization of \algname. 
Specifically, we consider three consistency functions: \algname-\textsc{RBF} with RBF kernel $f_{\text{rbf}}(\bm{\delta}) = \exp(-\bm{\delta} / \sigma)$, \algname-\textsc{pow} with power function $f_{\text{pow}}(\bm{\delta}) = \alpha(1 - \bm{\delta})^\beta$, and \algname-\textsc{sig} with sigmoid function $f_{\text{sig}}(\bm{\delta}) = 1 - \text{Sigmoid}(\gamma(\bm{\delta} - 0.5))$.
Experiments are conducted on the \texttt{SAMSum} dataset, with results presented in Figure~\ref{fig:exp-func}.

In general, different consistency score designs consistently enhance ensemble methods, demonstrating the broad applicability of \algname.
On average, \algname-\text{Pow} achieves the best performance with a 2.76 gain, followed by \algname-\text{RBF} (2.35) and \algname-\text{Sigmoid} (2.06).
The performance differences stem from how each function maps token disparity $\bm{\delta}$ to consistency weights.
RBF sharply favors well-aligned tokens, offering stable but conservative gains.
Power applies smoother decay, better balancing selectivity and inclusiveness, thus achieving the highest gain.
Sigmoid filters noise via soft thresholding but its limited smoothness dampens overall improvement.

\subsubsection{Case Study}

\begin{tcolorbox}
[left=3pt,right=3pt,top=3pt,bottom=3pt,colback=gray!5!white,colframe=black!75!white,title=Example 1]
\small
\textbf{Question:} what does the adrenal gland produce that is necessary for the sympathetic nervous system to function? \\
\textbf{Gold Answer:} epinephrine \\
\textbf{\texttt{OpenChat} Response:} \textcolor{red}{adrenaline} \xmark \\
\textbf{\texttt{InternLM} Response:} \textcolor{red}{epinephrine and norepinephrine} \xmark \\
\textbf{Vanilla Response:} \textcolor{red}{epineph\underline{ r}ine} \xmark \\
\textbf{\algname\ Response}: \textcolor{teal}{epineph\underline{r}ine} \cmark
\end{tcolorbox}

\begin{figure}[h]
    \centering
    \includegraphics[width=\linewidth, trim=6 0 8 0, clip]{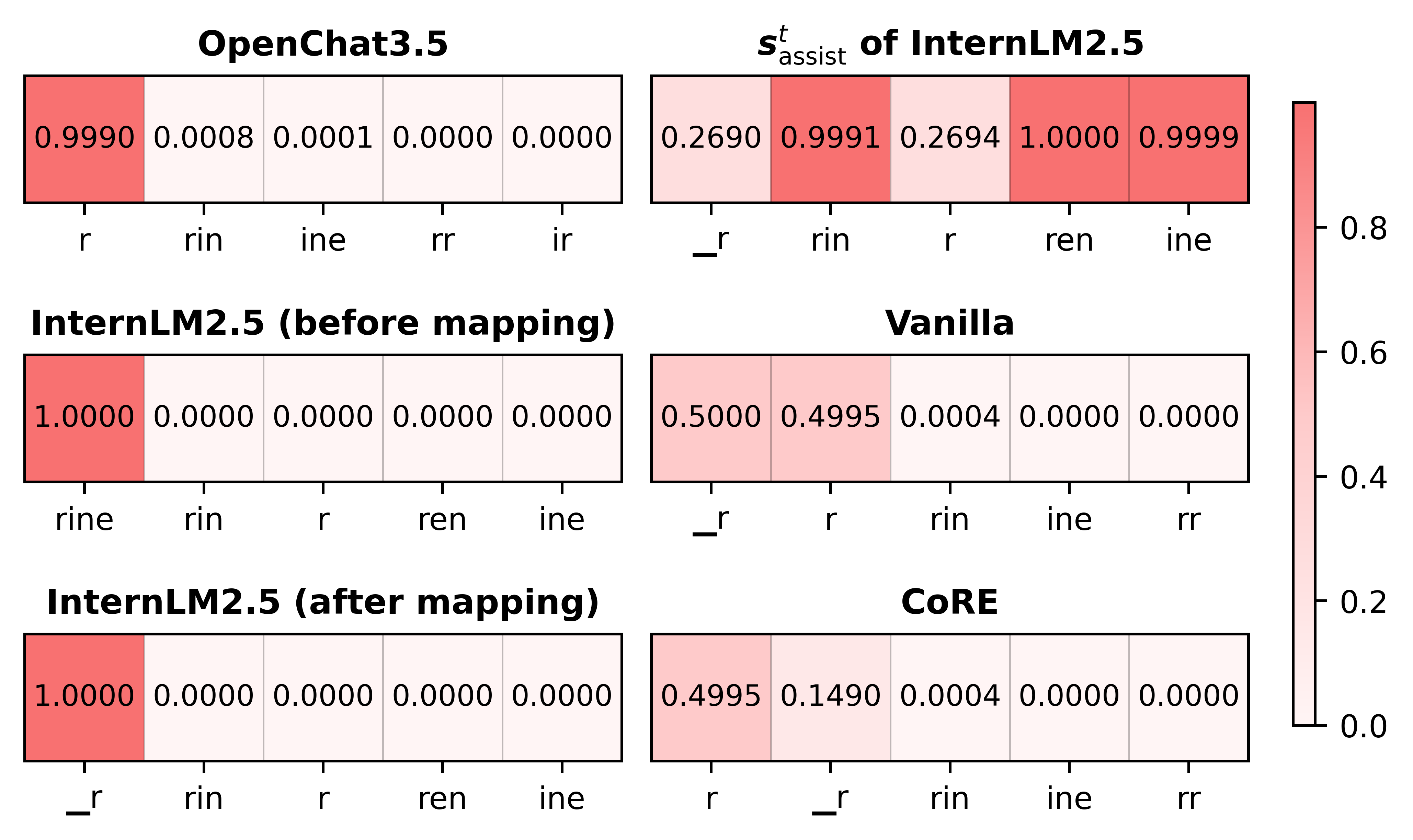}
    \vspace{-20pt}
    \caption{The Top-5 next token probability and the corresponding token consistency $\tcs_{\text{assist}}$ given generated text '\texttt{epineph}' in Example 1.}
    \vspace{-15pt}
    \label{fig:case}
\end{figure}

We present a case study on ensembling \texttt{OpenChat3.5} and \texttt{InternLM2.5} offering an intuitive illustration of how \algname\ operates.
As shown in Example 1 and Figure \ref{fig:case}, both individual models and vanilla ensemble fail to generate the correct answer, whereas augmenting with \algname\ yields the correct response.
The key difference lies in the fourth token following the generated text “\texttt{epineph}”: the vanilla ensemble predicts \codeinline{\_r}, while \algname\ correctly predicts \codeinline{r}.
Analyzing the token probabilities of \texttt{InternLM2.5} \textit{before} and \textit{after} token mapping reveals that the correct token \codeinline{rine} is incorrectly aligned to \codeinline{\_r}, leading the vanilla ensemble to an erroneous fusion.
In contrast, \algname\ identifies the misaligned token \codeinline{\_r} given its low consistency and downweights its influence, thereby penalizing unreliable tokens and models to produce the correct output.
More case studies are provided in Appendix~\ref{app:exp-case}.

%% file: sections/tab_benchmark.tex
\begin{table*}[]
\resizebox{\linewidth}{!}{
\begin{tabular}{cc|cc|cc|cc|cc|cc|cc|cc}
\toprule
\multicolumn{2}{c|}{\multirow{2}{*}{Method}}  & \multicolumn{2}{c|}{\textbf{GSM8K}} & \multicolumn{2}{c|}{\textbf{PIQA}} & \multicolumn{2}{c|}{\textbf{SAMSum}} & \multicolumn{2}{c|}{\textbf{TriviaQA}} & \multicolumn{2}{c|}{\textbf{NQ}} & \multicolumn{2}{c|}{\textbf{MMLU}} & \multicolumn{2}{c}{\textbf{Average}} \\
                        &                          & Top-2  & Top-3 & Top-2           & Top-3           & Top-2            & Top-3            & Top-2             & Top-3             & Top-2          & Top-3          & Top-2           & Top-3           & Top-2             & Top-3            \\ \midrule
\multirow{3}{*}{\rotatebox{90}{\textsc{MinED}}}  & Vanilla                  & 79.51  & 81.65                          & 85.47           & 82.32           & 43.79            & 42.18            & 67.50             & 59.63             & 25.07          & 22.22          & 68.38           & 66.66           & 61.62             & 59.11            \\
                        & \algname  & 82.41  & 83.78 & 85.75           & 87.49           & 49.25            & 49.55            & 72.17             & 72.85             & 29.83          & 29.61          & 68.42           & 69.77           & 64.64             & 65.51            \\
                        & $\Delta$    & \bluecell +2.90   &  \bluecell +2.13                         & \bluecell +0.28            & \bluecell +5.17            & \bluecell +5.46             & \bluecell +7.37             & \bluecell +4.67              & \bluecell +13.22             & \bluecell +4.76           & \bluecell +7.39           & \bluecell +0.04            & \bluecell +3.11            & \bluecell +3.02              & \bluecell +6.40            \\ \midrule
\multirow{3}{*}{\rotatebox{90}{\textsc{UniTE}}}  & Vanilla                  & 81.26  & 82.79                          & 85.64           & 86.89           & 48.03            & 47.99            & 70.93             & 64.90             & 28.17          & 25.84          & 68.61           & 69.56           & 63.78             &  63.00           \\
                        & \algname  & 82.71  & 83.70 & 85.96           & 87.38           & 49.41            & 49.77            & 73.41             & 73.61             & 30.39          & 29.94          & 68.63           & 69.91           & 65.09             & 65.72            \\
                        & $\Delta$    & \bluecell +1.45   & \bluecell +0.91                          & \bluecell +0.32            & \bluecell +0.49            & \bluecell +1.38             & \bluecell +1.78             & \bluecell +2.48              & \bluecell +8.71              & \bluecell +2.22           & \bluecell +4.10           & \bluecell +0.02            & \bluecell +0.35            & \bluecell +1.31              & \bluecell +2.72             \\ \midrule
\multirow{3}{*}{\rotatebox{90}{\textsc{EVA}}}    & Vanilla                  & 82.09  &  83.02                         & 87.81           & 87.43           & 48.16            & 47.83            & 73.47             & 73.02             & 30.00          & 29.09          & 70.62           & 71.07           & 65.36             & 65.24            \\
                        & \algname  & 83.40  & 83.24 & 87.81           & 87.76           & 49.43            & 50.27            & 73.13             & 73.23             & 29.53          & 29.09          & 70.62           & 70.91           & 65.65             & 65.75            \\
                        & $\Delta$    & \bluecell +1.31   & \bluecell +0.22                          & \bluecell +0.00            & \bluecell +0.33            & \bluecell +1.27             & \bluecell +2.44             & \redcell -0.34             & \bluecell +0.21              & \redcell -0.47          & \bluecell +0.00           & \bluecell +0.00            & \redcell -0.16           & \bluecell +0.29              & \bluecell +0.51             \\ \midrule
\multirow{3}{*}{\rotatebox{90}{\textsc{GaC}}}    & Vanilla                  & 80.74  & 83.02                          & 85.67           & 86.89           & 48.08            & 47.63            & 73.67             & 71.07             & 29.61          & 27.15          & 68.61           & 69.56           & 64.40             & 64.22            \\
                        & \algname  & 82.49  & 84.00 & 85.75           & 87.32           & 49.41            & 49.73            & 73.72             & 73.97             & 30.22          & 29.81          & 68.63           & 69.96           & 65.04             & 65.80            \\
                        & $\Delta$    & \bluecell +1.75   & \bluecell +0.98                          & \bluecell +0.08            & \bluecell +0.43            & \bluecell +1.33             & \bluecell +2.10             & \bluecell +0.05              & \bluecell +2.90              & \bluecell +0.61           & \bluecell +2.66           & \bluecell +0.02            & \bluecell +0.40            & \bluecell +0.64              & \bluecell +1.58   \\ \bottomrule         
\end{tabular}
}
\caption{Benchmark results. We report ensemble performance using the Top-2 and Top-3 base models on each dataset, with (\algname) and without (Vanilla) our method. The $\Delta$ rows report the performance gain from applying \algname, with \textcolor{blue}{blue} cells indicating improvement and \textcolor{red}{red} cells indicating degradation.}
\label{tab:benchmark}
\end{table*}

%% file: sections/tab_base_model.tex
\begin{table}[t]
\setlength{\tabcolsep}{4pt} 
\resizebox{\columnwidth}{!}{
\begin{tabular}{lcccccc} \toprule
\textbf{Model}      & \textbf{GSM8K} & \textbf{PIQA} & \textbf{SAMSum} & \textbf{TriviaQA} & \textbf{NQ} & \textbf{MMLU} \\ \midrule
\texttt{Llama3}     & \redcell 74.91 & \redcell 76.10 & 43.57 & \redcell 63.08 & \redcell 22.13 & \redcell 63.49 \\
\texttt{Mistral7b}  & 41.55 & 62.62 & \yellowcell 44.80 & 52.47 & 15.01 & 52.17 \\
\texttt{Qwen2.5}    & 36.76 & \yellowcell 80.03 & \redcell 43.77 & 43.50 & 12.96 & \yellowcell 64.80 \\
\texttt{InternLM2.5}& \bluecell 82.69 & \bluecell 86.91 & 42.21 & \yellowcell 63.23 & \yellowcell 26.62 & \bluecell 71.27 \\
\texttt{OpenChat3.5}& \yellowcell 76.35 & 62.62 & \bluecell 50.05 & \bluecell 72.13 & \bluecell 28.92 & 62.11 \\ \bottomrule
\end{tabular}
}
\caption{Base model performance. We use Blue, Yellow and Red to denote \textcolor{blue}{Top-1}, \textcolor{orange}{Top-2} and \textcolor{red}{Top-3} models.}
\vspace{-15pt}
\label{tab:base}
\end{table}

%% file: sections/tab_benchmark_deepen.tex
\begin{table}[]
\setlength{\tabcolsep}{4pt} 
\resizebox{\columnwidth}{!}{
\begin{tabular}{c|cc|cc|cc|cc}
\toprule
\multirow{2}{*}{\textbf{Method}} & \multicolumn{2}{c|}{\textbf{PIQA}} & \multicolumn{2}{c|}{\textbf{NQ}} & \multicolumn{2}{c|}{\textbf{MMLU}} & \multicolumn{2}{c}{\textbf{Average}} \\
                         & Top-2           & Top-3           & Top-2          & Top-3          & Top-2           & Top-3           & Top-2             & Top-3            \\ \midrule
Vanilla                  & 87.52           & 87.35           & 28.03          & 29.23          & 70.56           & 70.49           & 62.04             & 62.36            \\
\algname  & 87.74           & 87.35           & 28.14          & 29.53          & 70.57           & 70.5            & 62.15             & 62.46            \\
$\Delta$    & \bluecell +0.22            & \bluecell +0.00            & \bluecell +0.11           & \bluecell +0.30           & \bluecell +0.01            & \bluecell +0.01            & \bluecell +0.11              & \bluecell +0.10    \\ \bottomrule  
\end{tabular}
}
\caption{Ensemble performance with \textsc{DeePEn}.}
\vspace{-15pt}
\label{tab:benchmark-deepen}
\end{table}

%% file: sections/5-con.tex
\section{Conclusion}
In this paper, we study the robustness of LLM ensembles, a critical yet underexplored aspect in prior work.
Our analysis shows that ensemble failures often arise from inconsistencies at both the token and model levels.
To address this, we introduce \algname, a lightweight and plug-and-play framework that enforces consistency across multiple dimensions without incurring additional inference cost.
Extensive experiments show that \algname\ significantly enhances both the performance and robustness of existing ensemble methods.
Our findings highlight consistency as a key principle for building reliable LLM ensembles and open new directions for robustness-oriented ensemble in future research.

\section*{Limitations}
\algname\ requires access to token‐level logits of ensembled LLMs to enforce consistency at both token and model level, preventing its use with closed‐source or black‐box LLM APIs. 
Moreover, determining \emph{when to ensemble} remains an open question: if the main model is already confident or the assistant model exhibits low confidence, skipping ensembling may prevent unreliable outputs from degrading performance.
Finally, identifying \emph{which models to ensemble} is also worth exploring, while our model weights provide a soft balancing mechanism, future work could study more principled criteria for selecting beneficial model combinations.

\section*{Acknowledgements}
This work is supported by NSF (2433308) and
AFOSR (FA9550-24-1-0002).
The content of the information in this document does not necessarily reflect the position or the policy of the Government, and no official endorsement should be inferred.  The U.S. Government is authorized to reproduce and distribute reprints for Government purposes notwithstanding any copyright notation here on.

%% file: sections/app.tex
\appendix
\section*{Appendix}

\section{Experiments}~\label{app:exp-detail}

\vspace{-20pt}
\subsection{Examples of Token Misalignment}
We provide intuitive examples of token misalignment caused by different ensemble methods in Table~\ref{tab:token-mis}, validating that existing ensemble method may generate suboptimal or even irrelevant token mappings. 
Capable of handling token misalignment, \algname\ consistently improves the performance of existing ensemble methods.

\begin{table}[h]
\centering
\caption{Examples of misaligned tokens from source LLM (\texttt{OpenChat}) to mapped LLM (\texttt{InternLM}).}
\vspace{-5pt}
\begin{tabular}{ccc}
\toprule
\textbf{Method}  & \textbf{Source Token} & \textbf{Mapped Token} \\
\midrule
\textsc{MinED}   & \texttt{'riv'}      & \texttt{'{\_}IV'} \\
\textsc{UniTE}   & \texttt{'{\_}ind'}  & \texttt{'indows'} \\
\textsc{EVA}     & \texttt{'equation'} & \texttt{'align'} \\
\bottomrule
\end{tabular}
\label{tab:token-mis}
\end{table}

\subsection{Adapting \algname\ to \textsc{DeePEn}}
To adapt \algname\ for \textsc{DeePEn} that ensembles models in a latent embedding space, we normalize the unified token embeddings and transform them into probability distributions which \algname\ can operate on before ensemble. 
This adaptation is natural because \textsc{DeePEn}, rather than projecting tokens of assist LLMs into the vocabulary space of the main LLM, essentially projects tokens of ensembled LLMs into a joint latent embeddings space on top of common tokens across different LLMs. 
By interpreting the normalized embeddings as token probability vectors, \algname\ can be integrated seamlessly into embedding-based ensemble methods such as \textsc{DeePEn}.

\section{Additional Studies}\label{app:exp-case}
\subsection{Observation on Model Consistency}
\begin{figure}[ht]
    \centering
    \includegraphics[width=\linewidth]{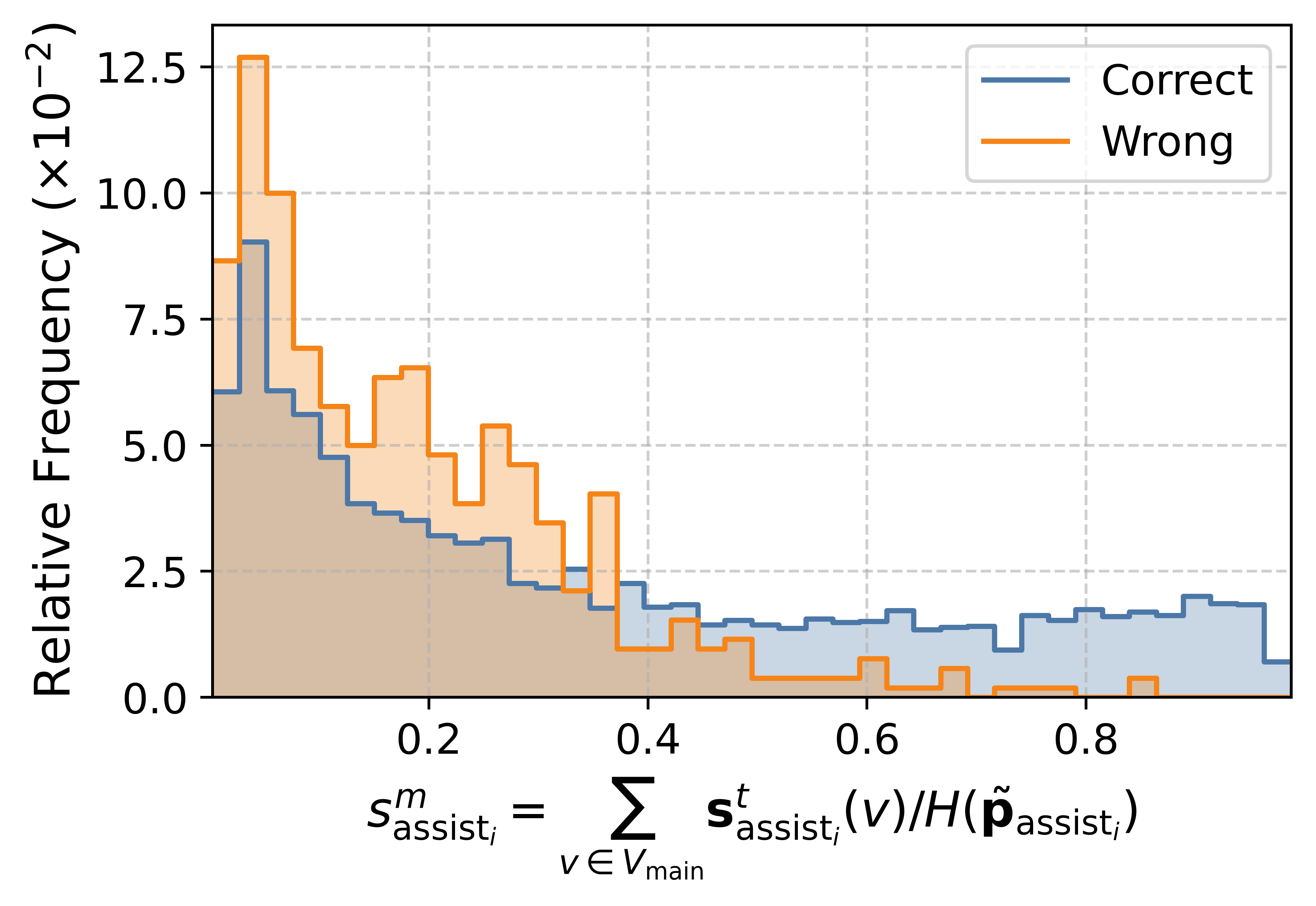}
    \caption{Model score visualization.}
    \label{fig:model-score}
\end{figure}

In addition to observations in~\ref{sec:obs}, we directly evaluate how model score, measured by the quotient of model consistency and confidence, i.e., $\mcs_{\text{assist}_i}=\sum_{v\in V_{\text{main}}}{\tcs_{\text{assist}_i}(v)/H(\tp_{\text{assist}_i})}$, indicates the correctness of responses. As shown in Figure~\ref{fig:model-score}, correct answers exhibit higher model score than the wrong ones. A one-sided test of $H_0: \mathbb{E}[\mcs_{\text{assist}_i} | \text{wrong}] \geq \mathbb{E}[\mcs_{\text{assist}_i} | \text{correct}]$ yields a $p$-value of $2.5\times 10^{-79}$, confirming that correct answers have larger model scores than the wrong ones. We summarize this into Observation~\ref{obs:model-score}, which further validates our design of the model score.

\begin{observation}[Model Score]\label{obs:model-score}
    Large model score measured by the quotient of model consistency and confidence signals correct answer.
\end{observation}

\subsection{Additional Case Studies}
We present additional case studies on the Comprehensive Examination dataset (\texttt{MMLU}), Summarization dataset (\texttt{SAMSum}), and Reasoning dataset (\texttt{GSM8K}) to provide a more intuitive understanding of how \algname\ operates.
As illustrated in Figure~\ref{fig:add_case}, correct answers are highlighted in \textcolor{teal}{green} and incorrect ones in \textcolor{red}{red}.
\algname\ successfully produces the correct answers where both individual base models and the vanilla ensemble fail.
\input{sections/tab_app_case}

\section{Datasets}\label{sec:app-dataset}
\noindent\textbf{\texttt{GSM8K}}~\cite{cobbe2021training} is a multi-step arithmetic reasoning dataset consisting of 1,319 linguistically diverse grade-school math word problems authored by humans. Each problem requires explicit intermediate reasoning to reach a numeric answer. Models are prompted with four chain-of-thought exemplars, and the final prediction is deemed correct only if the predicted numeric answer exactly matches the gold label. 
We use accuracy as the evaluation metric.

\noindent\textbf{\texttt{PIQA}}~\cite{bisk2020piqa} is a benchmark for physical commonsense reasoning. Each of its 1,838 examples presents a naturalistic physical situation and two possible solutions, requiring models to choose the more plausible one. It evaluates a model’s ability to reason about everyday physical interactions. We use exact match as the evaluation metric.

\noindent\textbf{\texttt{SAMSum}}~\cite{gliwa2019samsum} is a dialogue summarization dataset composed of multi-turn conversations between fictitious participants. The task requires models to generate concise and coherent summaries that capture key information. No in-context examples are used. We use rouge-1 score as the evaluation metric.

\noindent\textbf{\texttt{TriviaQA}}~\cite{joshi2017triviaqa} contains 11,313 open-domain factoid questions authored by trivia enthusiasts, paired with Wikipedia-based ground-truth answers. Following prior work, each example includes five in-context QA exemplars, and model accuracy is measured by exact match between predicted and reference answers. We use exact match as the evaluation metric.

\noindent\textbf{\texttt{NQ}}~\cite{kwiatkowski2019natural} consists of 3,610 real anonymized Google search queries paired with short answers from Wikipedia articles. We follow prior work using 5-shot in-context prompting. We use exact match as the evaluation metric.

\noindent\textbf{\texttt{MMLU}}~\cite{hendrycks2020measuring} (Massive Multitask Language Understanding) is a 57-subject multiple-choice benchmark spanning STEM, humanities, and social sciences. Each question has four answer choices, and models are evaluated under 5-shot settings with 5,000 test examples. We use exact match as the evaluation metric.

\section{More Related Works}
\label{app:more_related_works}

\paragraph{Pre-trained Foundation Models.}
The landscape of NLP has been revolutionized by the emergence of LLMs trained on massive corpora~\cite{achiam2023gpt,touvron2023llama,jiang2023mistral7b}, with wide applications in reasoning~\cite{lin2024duquant,lin2025quantization}, language understanding~\cite{team2023gemini,team2024qwen2}, and many more~\cite{li2025language, li2025can}.
However, due to differences in training data, architectures, and tokenization schemes, these models exhibit diverse strengths and weaknesses across different tasks~\cite{wang2023openchat,team2023internlm}. 
For instance, some models may excel at mathematical reasoning~\cite{guo2025deepseek,team2023gemini} while others specialize in dialogue or summarization~\cite{achiam2023gpt}.
The heterogeneity among foundation models serves as the primary motivation for test-time ensemble strategies, which aim to integrate their complementary capabilities to achieve superior performance and reliability.

\paragraph{Model Ensemble and Alignment.}
In the era of big data and AI, machine learning models are inherently heterogeneous~\cite{zheng2024heterogeneous,zeng2025hierarchical,zeng2025interformer}, characterized by diverse model architectures~\cite{hochreiter1997long,vaswani2017attention,dosovitskiy2020image}, data modalities~\cite{yan2021bright,yan2021dynamic,yan2022dissecting,zeng2023generative,zeng2024graph,zeng2025pave}, and distinct tokenization frameworks~\cite{sennrich2016neural,kudo2018sentencepiece,wu2016google}.
Model ensemble is a well-established paradigm to combine various model capabilities for improved performance, with wide applications in multi-modal learning~\cite{zeng2026subspace,ning2025graph4mm}, language models~\cite{huang2024ensemble,yao2024determine,yu2024breaking}, time-series analysis~\cite{liu2025breaking,lin2024backtime,lin2025cats}, recommendation~\cite{liu2024collaborative,liang2025external,yoo2024ensuring}, social analysis~\cite{xu2024slog,yan2024pacer,yan2024thegcn,yan2025answer} and many more.
A prerequisite for ensembling heterogeneous models is to bridge their gaps via alignment~\cite{zeng2023parrot,zeng2024hierarchical,yu2025joint,yu2025planetalign}.
Specifically, token alignment ensures that probability distributions from heterogeneous LLMs are mapped onto a unified space before fusion.
Early approaches align token spaces via minimum edit distance~\cite{fu2023specializing,wan2024knowledge,mavromatis2024pack}, which captures structural differences but often incurs high computational overhead. 
Subsequent methods focus on vocabulary mapping in the text space~\cite{yu2024breaking,yao2024determine} by constructing mappings based on exact or prefix matches between token strings. 
Recent works utilize token embeddings to exploit semantic information that text matching may fail to capture~\cite{huang2024ensemble,xu2024bridging}, learning projection functions or cross-model embeddings to transform heterogeneous token distributions into a shared representation space.

\paragraph{LLM Robustness.}
Ensuring model robustness is critical for LLM deployment in real-world applications. 
Existing research largely focuses on safety against adversarial attacks and stability under distribution shifts. 
Adversarial robustness studies show that LLMs are vulnerable to jailbreak prompts~\cite{wei2023jailbroken,lin2026alert} or perturbations that trigger unintended behaviors~\cite{zou2023universal}. 
Meanwhile, out-of-distribution (OOD) robustness aims to maintain performance amidst noisy inputs or domain shifts~\cite{zhu2023prompt}.
Common mitigation techniques range from safety alignment (RLHF)~\cite{bai2022training,zhang2025improving} to test-time interventions like uncertainty quantification~\cite{kadavath2022language,lin2023generating}. 
While most existing works focus on individual model, we study the multi-model setting, showing that consistency constraints can serve as a robustification mechanism to mitigate the impact of unreliable signals within an ensemble.

\section{Potential Risks}
The proposed \algname, while enhancing robustness and reliability through token- and model-level consistency evaluation, also introduce potential risks that warrant careful consideration.
\algname\ relies on access to fine-grained token probability distributions from individual models to compute consistency scores, which limits its applicability to open-weight or transparent systems. When applied to closed-source or API-based models, this requirement may lead to implementation incompatibility or potential breaches of usage policies.

\section{Use Or Create Scientific Artifacts}  
Our work is built on public benchmarks and contributes new code resources to the community. For evaluation, we use widely adopted datasets including \texttt{GSM8K}, \texttt{PIQA}, \texttt{SAMSum}, \texttt{TriviaQA}, \texttt{NQ}, and \texttt{MMLU}, without making any changes to their original content. In addition, we introduce and release the codebase of \algname\ at \url{https://github.com/zhichenz98/CoRE-EACL26}, providing a clear and well-structured repository to improve the accessibility of our research.

\subsection{Cite Creators Of Artifacts}
All external artifacts are properly credited to their original publications and repositories.

All benchmarks used in this work, including \texttt{GSM8K}~\cite{cobbe2021training}, \texttt{PIQA}~\cite{bisk2020piqa}, \texttt{SAMSum}~\cite{gliwa2019samsum}, \texttt{TriviaQA}~\cite{joshi2017triviaqa}, \texttt{NaturalQuestions}~\cite{kwiatkowski2019natural}, and \texttt{MMLU}~\cite{hendrycks2009measuring}, are credited to their respective authors.

Each LLM model used in this work, including \texttt{Llama-3-8B-Instruct}~\cite{dubey2024llama}, \texttt{Mistral-7B-Instruct-v0.1}~\cite{jiang2023mistral7b}, \texttt{Qwen2.5-3b-Instruct}~\cite{team2024qwen2}, \texttt{InternLM2.5-7b-Chat}~\cite{team2023internlm} and \texttt{openchat-3.5-0106}~\cite{wang2023openchat}, is referenced through its official technical report or HuggingFace model card to ensure appropriate acknowledgment of all upstream contributions.

\subsection{Discuss The License For Artifacts}
We comply with the licenses of all artifacts used or released in this work. The benchmarks are distributed under Creative Commons or public-domain terms, following the conditions specified by their original maintainers.
Model checkpoints retain their original open-source licenses, and our own code and generated data are released under the MIT license. We will release our code upon publication.

\subsection{Artifact Use Consistent With Intended Use}
We confirm that our use of datasets and pre-trained models is consistent with their intended purpose. Benchmarks are used solely for inference and evaluation, which aligns with their terms of service, and no modified model weights are redistributed. The released code is for inference only and does not support fine-tuning or commercial redistribution of the original checkpoints.

\subsection{Documentation Of Artifacts}
We provide detailed documentation of all evaluation datasets and model combinations used in our experiments. Specifically, Appendix~\ref{sec:app-dataset} describes the coverage, task type, and evaluation metric for each dataset (e.g., reasoning, summarization, and knowledge QA), while Appendix~\ref{sec:app-exp} the LLMs included in the ensemble along with their sizes and sources. Together, these materials ensure transparency regarding the domains, data characteristics, and model diversity involved in our study.

\section{Statistics For Data}

\subsection{GSM8K \textnormal{(Arithmetic Reasoning)}}
\begin{itemize}[noitemsep, topsep=0pt]
  \item Number of entries: 1319
  \item Average question length: 239.9 characters
  \item Average answer length: 272.3 characters
\end{itemize}

\subsection{PIQA \textnormal{(Commonsense Reasoning)}}
\begin{itemize}[noitemsep, topsep=0pt]
  \item Number of entries: 1838
  \item Average question length: 36.1 characters
  \item Average answer length: 1.0 character
\end{itemize}

\subsection{SAMSum \textnormal{(Summarization)}}
\begin{itemize}[noitemsep, topsep=0pt]
  \item Number of entries: 819
  \item Average question length: 521.6 characters
  \item Average answer length: 108.8 characters
\end{itemize}

\subsection{TriviaQA \textnormal{(Knowledge)}}
\begin{itemize}[noitemsep, topsep=0pt]
  \item Number of entries: 6000
  \item Average question length: 78.8 characters
  \item Average answer length: 267.9 characters
\end{itemize}

\subsection{NaturalQuestions \textnormal{(Knowledge)}}
\begin{itemize}[noitemsep, topsep=0pt]
  \item Number of entries: 3610
  \item Average question length: 47.7 characters
  \item Average answer length: 24.7 characters
\end{itemize}

\subsection{MMLU \textnormal{(Comprehensive Examination)}}
\begin{itemize}[noitemsep, topsep=0pt]
  \item Number of entries: 14042
  \item Average question length: 274.5 characters
  \item Average answer length: 1.0 character
\end{itemize}

\section{Computational Experiments}\label{sec:app-exp}
All computational experiments in this work are fully reproducible, with details provided in the main text and the Appendix.

\subsection{Model Size And Budget}
For each LLM used, we specify its total parameter count as follows: Llama-3-8B-Instruct contains around 8 billion parameters; Mistral-7B-Instruct-v0.1, InternLM2.5-7b-Chat, and openchat-3.5-0106 contains around 7 billion parameters; Qwen2.5-3b-Instruct contains around 3 billion parameters. The total compute budget for all experiments is approximately 500 A100 GPU hours.

\subsection{Experiment Setup And Hyperparameters}
We describe experimental settings for all experiments in Section~\ref{sec:exp-setup}.
For hyperparameter settings, we set $\sigma=0.5$ in the RBF kernel, $\alpha=1.0, \beta=1.0$ in the power function, and $\gamma=1.0$ in the sigmoid function.
We clip the normalized model weight for the main LLM to be at least 0.5 to ensure its major contribution to the ensemble results.

\subsection{Descriptive Statistics}
For each result in the main text and Appendix, we report the mean across multiple runs.

\subsection{Parameters For Packages}
The existing packages used are specified as follows. We use PyTorch (v2.8.0) as the core deep learning framework, together with HuggingFace Transformers (v4.56.1) and Tokenizers (v0.22.0) for model implementation and text pre-processing. For efficient multi‑GPU and distributed training we rely on Accelerate (v1.10.1).

%% file: sections/tab_app_case.tex
\begin{figure*}[t]
\centering

\begin{tcolorbox}
[left=3pt,right=3pt,top=3pt,bottom=3pt,colback=gray!5!white,colframe=black!75!white,title=Example 2 (Comprehensive Examination)]
\small
\textbf{Question:} A person weighs 62 kg. Their drug dose is 15 mg/kg. How many grams is their dose? Choose one answer from the following: \\
A. 930~~B. 93~~C. 9.3~~D. 0.93 \\
\textbf{Gold Answer:} D \\
\textbf{\texttt{InternLM} Response:} \textcolor{red}{C} \xmark \\
\textbf{\texttt{Qwen} Response:} \textcolor{red}{C} \xmark \\
\textbf{Vanilla Response:} \textcolor{red}{C} \xmark \\
\textbf{\algname\ Response:} \textcolor{teal}{D} \cmark
\end{tcolorbox}

\begin{tcolorbox}
[left=3pt,right=3pt,top=3pt,bottom=3pt,colback=gray!5!white,colframe=black!75!white, title=Example 3 (Summarization)]
\small
\textbf{Question:} Generate a concise summary for the chat: \\
Kelly: Should I put on the red dress? \\
Meghan: oh yes!!!! \\
Mary: I'll wear a red dress too, and bright, red lipstick \\
Kelly: bloody Mary! \\
Mary: hahahaha \\
\textbf{Gold Answer:} Kelly and Mary will wear red dresses. Mary will wear a red lipstick too. \\
\textbf{\texttt{OpenChat} Response:} Kelly, Meghan, and Mary all decide to wear red dresses and bright red lipstick. \textcolor{red}{(Rouge Score=0.55)} \\
\textbf{\texttt{Mistral} Response:} Kelly is considering wearing a red dress and Mary suggests she wear red lipstick to go with it. \textcolor{red}{(Rouge Score=0.64)} \\
\textbf{Vanilla Response:} Kelly is considering wearing a red dress and Mary suggests wearing a red dress and red lipstick. Kelly responds with "bloody Mary". \textcolor{red}{(Rouge Score=0.55)} \\
\textbf{\algname\ Response:} Kelly is considering wearing a red dress and Mary will also wear a red dress and red lipstick. \textcolor{teal}{(Rouge Score=0.73)}
\end{tcolorbox}

\begin{tcolorbox}
[left=3pt,right=3pt,top=3pt,bottom=3pt,colback=gray!5!white,colframe=black!75!white, title=Example 4 (Reasoning)]
\small
\textbf{Question:} John plans to sell all his toys and use the money to buy video games. He has 13 lego sets and he sells them for \$15 each. He ends up buying 8 video games for \$20 each and has \$5 left. How many lego sets does he still have? \\
\textbf{Gold Answer:} He spent \$160 on video games because 8 x 20 = 160 \\
He earned \$165 from his sale because 160 + 5 = 165 \\
He sold 11 lego sets because 11 x 15 = 165 \\
He has 2 sets left because 13 - 11 = 2 \\
Therefore, the answer is 2 \\
\textbf{\texttt{InternLM} Response:} John sells 13 lego sets for \$15 each, so he earns 13*15 = \$195 \\
He buys 8 video games for \$20 each, so he spends 8*20 = \$160 \\
He has \$5 left, so the total amount of money he earned is \$160 + \$5 = \$165 \\
Since he earned \$195 from selling his lego sets, the number of lego sets he still has is \$195 - \$165 = \$30 \\
The answer is \textcolor{red}{30 \xmark} \\
\textbf{\texttt{OpenChat} Response:} John sells his 13 lego sets for \$15 each, so he makes 13*\$15=\$195 \\
He uses this money to buy 8 video games for \$20 each, so he spends 8*\$20=\$160 \\
He has \$5 left, so he spent \$195-\$160-\$5=\$30 on lego sets \\
Since he sells each lego set for \$15, he sells 30/15=2 lego sets \\
He started with 13 lego sets and sold 2, so he still has 13-2=11 lego sets \\
The answer is \textcolor{red}{11 \xmark} \\
\textbf{Vanilla Response:} John sells 13 lego sets for \$15 each, so he earns 13*15 = \$195 \\
He buys 8 video games for \$20 each, so he spends 8*20 = \$160 \\
He has \$5 left, so he spent a total of 195-5 = \$190 on lego sets \\
Since he sold 13 lego sets for \$15 each, he must have had 190/15 = 13 lego sets \\
The answer is \textcolor{red}{13 \xmark} \\
\textbf{\algname\ Response:} John sells 13 lego sets for \$15 each, so he earns 13*15 = \$195 from selling the lego sets. \\
He buys 8 video games for \$20 each, so he spends 8*20 = \$160 on video games. \\
He has \$5 left, so he spent a total of 195-160-5 = \$30 on other things. \\
Since each lego set is sold for \$15, he still has 30/15 = 2 lego sets left. \\
The answer is \textcolor{teal}{2 \cmark}
\end{tcolorbox}

\caption{Additional case study across comprehensive examination, summarization, and reasoning benchmarks. Correct and incorrect responses are highlighted in \textcolor{teal}{green} and \textcolor{red}{red}, respectively.}
\label{fig:add_case}
\end{figure*}